%% file: main.tex
\documentclass[letterpaper]{article}
\usepackage[preprint]{aaai2027}
\usepackage[hyphens]{url}
\usepackage{graphicx}
\usepackage{natbib}
\usepackage{caption}
\usepackage{booktabs}
\usepackage{amsmath}
\usepackage{amssymb}
\usepackage{multirow}
\usepackage{makecell}
\usepackage{pifont}
\usepackage{placeins}
\usepackage{cuted}
\newcommand{\cmark}{\ding{51}}
\newcommand{\xmark}{\ding{55}}

\title{Knowledge before Reasoning: EC-Reason-Bench, a Training-Free Diagnostic Benchmark for LLM Enzyme Classification}
\author{
    Linyu Li\textsuperscript{\rm 1},
    Zhi Jin\textsuperscript{\rm 1,$\dagger$},
    Yichi Zhang\textsuperscript{\rm 2},
    Dongming Jin\textsuperscript{\rm 1},\\
    Yuanpeng He\textsuperscript{\rm 1},
    Huanyao Zhang\textsuperscript{\rm 1},
    Xuan Zhang\textsuperscript{\rm 1},\\
    Gadeng Luosang\textsuperscript{\rm 3},
    Nyima Tashi\textsuperscript{\rm 3}
}
\affiliations{
    \textsuperscript{\rm 1}School of Computer Science, Peking University, Beijing, China\\
    \textsuperscript{\rm 2}College of Computer Science and Technology, Zhejiang University, Hangzhou, China\\
    \textsuperscript{\rm 3}School of Computer Science, Tibet University, Lhasa, China\\
    \textsuperscript{$\dagger$}Corresponding author.\\
    linyuli@stu.pku.edu.cn, zhijin@pku.edu.cn
}

\begin{document}

\maketitle

\begin{abstract}
    Enzyme function prediction is a hierarchical, knowledge-intensive form of protein function classification. Existing benchmarks expose an anomaly: general LLMs often get the coarse first level right, yet once asked for a complete EC number their accuracy at levels two through four drops to almost zero, while specialized models and tools stay usable. We propose EC-Reason-Bench, a training-free, diagnostic evaluation protocol built to answer two questions: why general LLMs score close to nothing on EC number prediction, and how much of that loss can be recovered without updating a single weight. We break enzyme classification ability into four orthogonal levers that can each be measured on their own: output structure, external knowledge, reasoning structure, and reasoning robustness. We test each lever with an inference-time method against a shared zero-shot baseline reproducing previously reported near-zero performance. Experiments with several strong reasoning LLMs yield four main findings. First, external knowledge is decisive and must precede reasoning: uniformly low closed-book performance rises sharply with open-book access, narrowing model gaps. Second, in closed-book settings, whether cascading and chain-of-thought help or hurt depends on a model’s tendency to abstain. Third, once evidence is available the aggregate score of the best LLM setting is indistinguishable from simply voting the EC numbers of the nearest retrieved neighbors; that tie is an artifact of averaging, and it hides a large gain on adversarial evidence set against an equally large loss on multi-functional enzymes. Reasoning over evidence therefore acts as an arbiter of conflicting neighbors rather than as a source of knowledge, and no single-number leaderboard can see it. Fourth, accuracy obeys a law of homology availability.

\end{abstract}

\section{Introduction}
\begin{figure}[t]
    \centering
    \includegraphics[width=0.85\columnwidth]{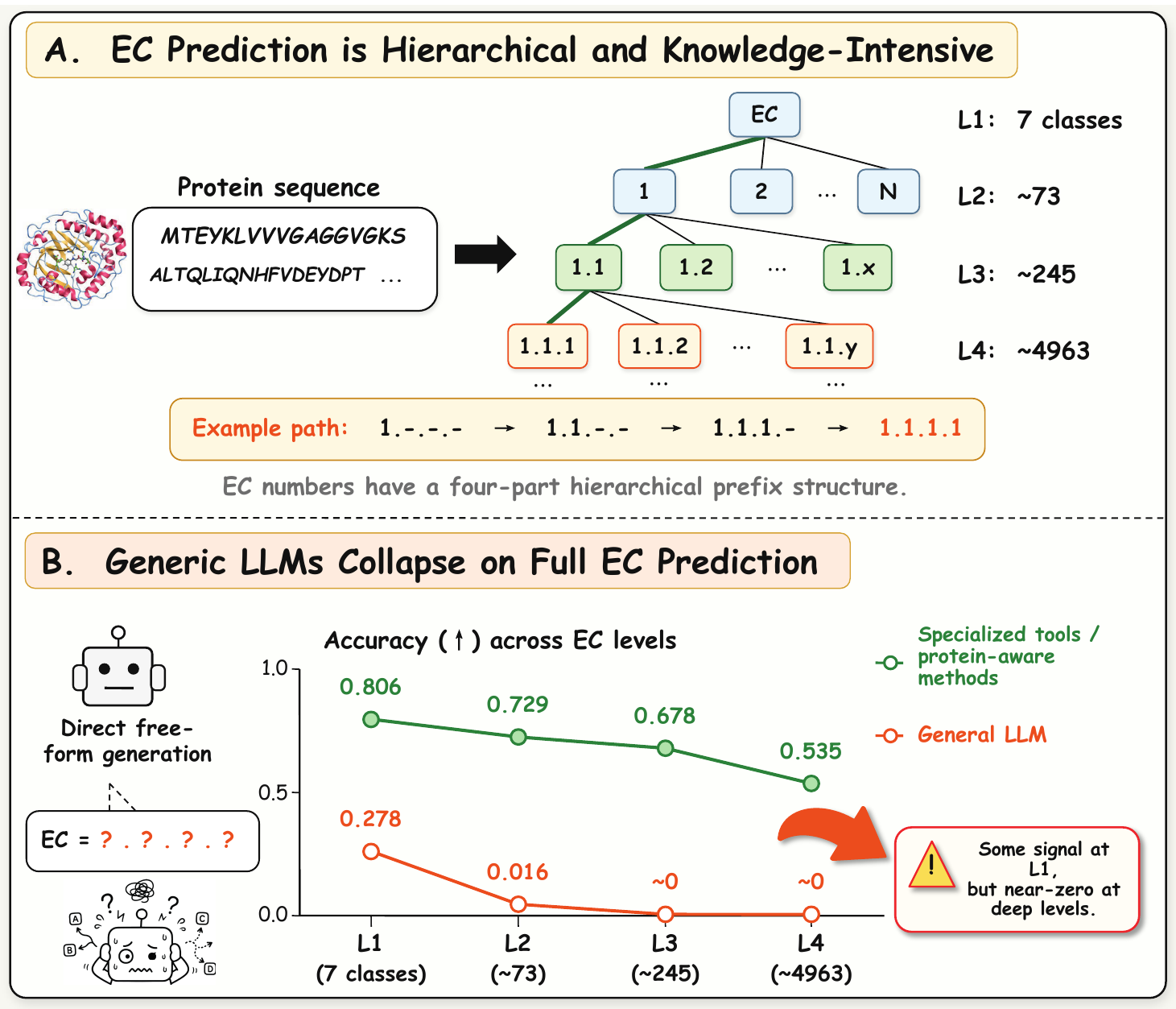}
    \caption{\textbf{Motivation for EC-Reason-Bench.} (a) The EC tree. (b) Level-wise accuracy under one-shot free generation: general LLMs collapse after the first level, while specialist and protein-aware methods decline smoothly.}
    \label{fig:intro}
    \end{figure}
Enzyme function annotation is the step that links sequence to metabolism, and it underpins pathway reconstruction, drug discovery, and protein engineering. Its standard form is the Enzyme Commission (EC) number. An EC number \texttt{a.b.c.d} describes the reaction an enzyme catalyzes from coarse to fine, and it naturally forms a four-level tree (Figure~\ref{fig:intro}(a)) with 7 classes, about 73 subclasses, about 245 sub-subclasses, and exactly 4963 valid leaves. A shared prefix is a parent-child relation, so this tree gives any classifier a strong structural prior for free.

\paragraph{The anomaly.} As Figure~\ref{fig:intro}(b) shows, general LLMs collapse on this task. They can still guess the 7-way first level with an accuracy of roughly $0.28$ to $0.54$, but once asked for a complete EC number their accuracy from the second level to the fourth falls to almost zero, while specialist approaches such as BLASTp homology search, Foldseek structure search, contrastive CLEAN and multi-modal hyperbolic models decline smoothly and stay usable. This is the fingerprint of formatting and hallucination, not of complete ignorance: a model that knew nothing would also be near chance at the first level, and dense protein representations restore accuracy at once. Existing benchmarks, however, report this near-zero number only as a token entry alongside specialist models, never asking where it comes from or how much can be recovered.

\paragraph{Two layers of loss.} We argue that the near-zero failure should be split into two parts and \emph{measured separately}. \emph{Formatting} is the fixable way the task and the evaluation are organized: handing a raw sequence to a text model, generating once over a closed vocabulary of 4963 classes, ignoring the hierarchy, forbidding abstention, and scoring multi-label enzymes with a mismatched rule. \emph{Knowledge} is a genuine gap: enzyme function depends on external biological priors such as homology, domain annotation, structure and active sites, so low-homology and long-tail samples remain hard even after every formatting problem is fixed. Existing benchmarks mix the two in one number.

\begin{figure*}[t]
\centering
\includegraphics[width=0.85\textwidth]{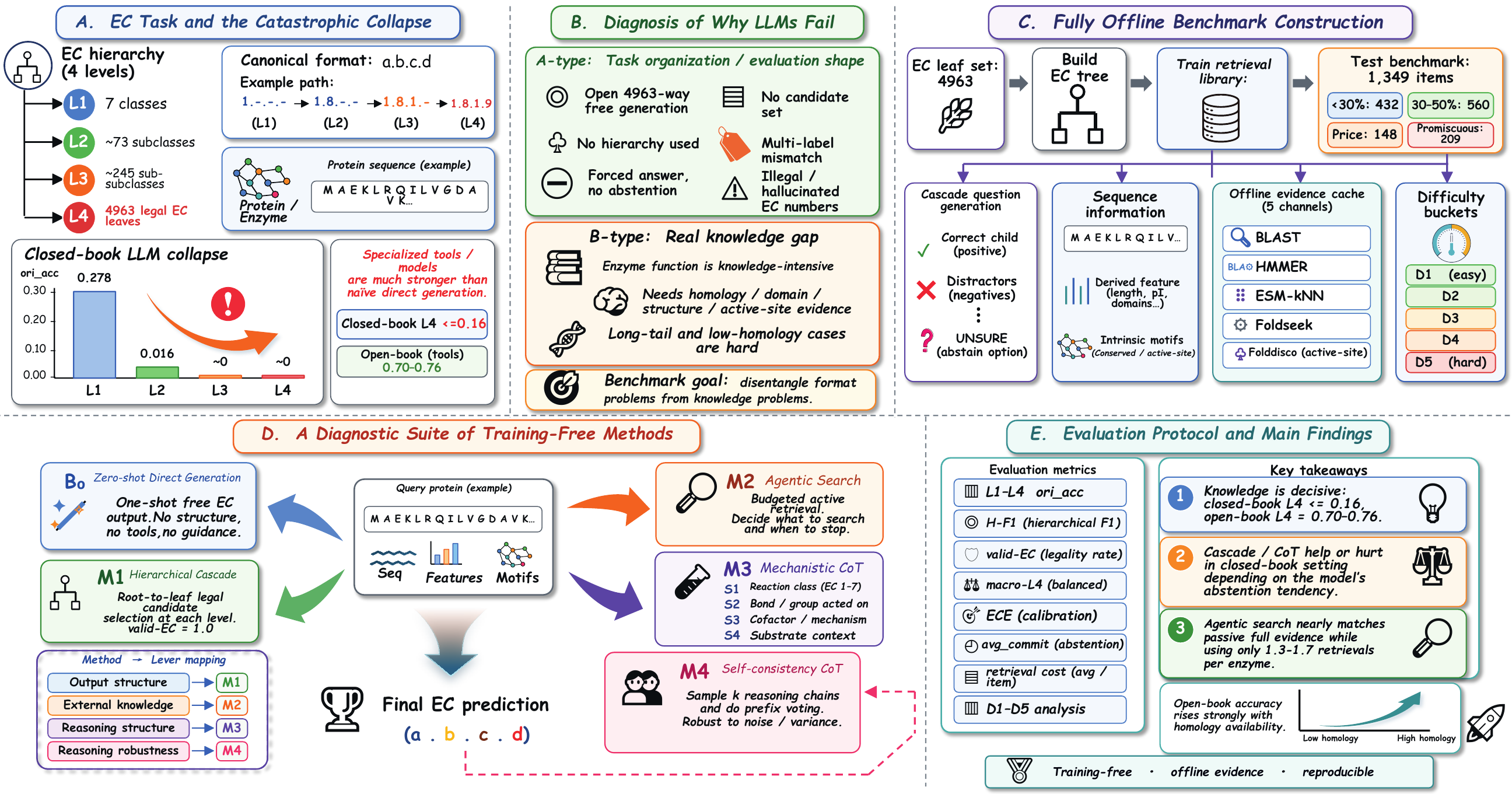}
\caption{\textbf{Overview of EC-Reason-Bench.} (A) Task and collapse. (B) Diagnosis of the near-zero failure into fixable formatting failures (A-type) and genuine knowledge failures (B-type). (C) Fully offline construction of the EC tree, the retrieval database and 1349 questions over four splits; \emph{domains} there means motifs scanned from the sequence, not database annotation. (D) The zero-shot baseline B0 and the four orthogonal levers M1--M4. (E) Evaluation protocol and main findings.}
\label{fig:model}
\end{figure*}

\paragraph{Our approach.} EC-Reason-Bench recasts this flat classification task into a diagnostic evaluation ablated along four \emph{orthogonal levers}: output structure, external knowledge, reasoning structure, and reasoning robustness. Each lever has one \emph{training-free} inference-time method, M1 to M4, compared against a single zero-shot baseline B0 that reproduces the reported catastrophic setting (Figure~\ref{fig:model}). The four methods share the same questions, the same offline evidence cache and the same hierarchical scorer, so they support clean ablation and can be stacked. Our contributions are as follows:
\par (1)  The first study that treats general LLMs as \emph{first-class subjects} in enzyme function classification, showing that their collapse is reversed \emph{with external knowledge alone, without changing any weight}: fourth-level accuracy is at most $0.16$ closed-book and rises to $0.70$--$0.76$ open-book.
\par (2)  A diagnostic benchmark built on four orthogonal levers that make formatting and knowledge separately measurable, each isolated by one training-free method over questions and evidence constructed fully offline under a strictly inductive setting.
\par (3)  A full run of nine settings on the complete set for five strong reasoning models, quantifying that knowledge is decisive, that format repair helps or hurts with a model's readiness to abstain, that accuracy obeys a law of homology availability, and that the LLM's net gain over neighbor voting is confined to the splits where evidence misleads or multiplies.

\section{Related Work}

\paragraph{Enzyme function prediction and benchmarks.} Specialist methods have long dominated EC prediction: sequence and structure homology search~\cite{altschul1990basic,potter2018hmmer,van2024fast}, contrastive learning~\cite{yu2023enzyme}, protein language models with a classification head~\cite{lin2023evolutionary,elnaggar2021prottrans,wang2025s}, hierarchical supervision~\cite{ryu2019deep,sanderson2023proteinfer}, and a recent turn toward structure and multiple modalities~\cite{van2025topec,yuan2025proteinf3s,pmlr-v267-zhang25cz,song2026improving}. HIT-EC~\cite{dumontet2026trustworthy} shows that hierarchical constraints and the willingness to answer are coupled in specialist models too, matching our M1 and M4. Among EC-specific benchmarks, CARE~\cite{yang2024care} first reported the catastrophic general-LLM baseline, EC-Bench~\cite{davoudi2026ec} compares four families of methods, and PoinnCARE~\cite{xie2026poinncare} sets the state of the art under an inductive retrieval setting. All three evaluate \emph{specialist} models and never ask why LLMs fail. The appendix extends this discussion and tabulates our positioning.

\paragraph{LLM reasoning and inference-time methods.} Pika~\cite{carrami2024pqa} shows that feeding ESM embeddings to an LLM works far better than prompting a general model~\cite{hurst2024gpt} with text alone. Closest to us is LLaPA~\cite{peng2024textttllapa}, which also starts from the collapse of direct EC generation but ties EC renumbering and dual-path retrieval into one \emph{trained} system, so it cannot say how much either part contributes, which is exactly what our levers separate. PFUA~\cite{fan2026interleaved} shows that text-only chains of thought do not transfer to protein function while tool use more than doubles accuracy, and Bio-KCoT~\cite{lyu2025knowledge} anchors them in a knowledge graph; together they set the motivation and the limits for M2 and M3. The levers themselves rest on established paradigms: least-to-most prompting~\cite{zhou2022least}, chain of thought~\cite{wei2022chain}, self-consistency~\cite{wang2022self}, ReAct~\cite{yao2022react}, retrieval-augmented generation~\cite{lewis2020retrieval}, training-free hierarchical classification~\cite{im2023hierarchical,bhambhoria2023simple} and SELF-DISCOVER~\cite{zhou2024self}. Our contribution is not a new paradigm but a controlled protocol that measures the \emph{net effect} of each on one knowledge-intensive task.

\section{EC-Reason-Bench: Task and Design}

\subsubsection{Task: EC Numbers Form a Tree}
Let the test set be $\{(x_i,Y_i)\}_{i=1}^{N}$, where $x_i$ is a query enzyme and $Y_i$ its ground-truth set of EC numbers, which may hold more than one label. An EC number $e=a.b.c.d$ is a root-to-leaf path in the EC tree and $\pi_\ell(e)$ its $\ell$-th level prefix, so that $\pi_2(1.1.5.4)=1.1$; the set of valid leaves satisfies $|\mathcal E|=4963$. Constraining predictions to this tree is where our benchmark departs from flat classification over 4963 classes.

\subsubsection{Diagnosis and Four Orthogonal Levers}
The formatting side of the failure breaks down into concrete, fixable choices: a raw sequence is close to noise for a text-only model, one-shot free generation over a closed vocabulary is bound to hallucinate invalid EC numbers, ignoring the hierarchy makes one early mistake fatal, forced answering leaves no room to abstain, and multi-label scoring is mismatched. The knowledge side is where external biological priors are missing, an unconstrained chain of thought degenerates into surface-level paraphrase, and long-tail or low-homology samples are hard in themselves. Against these we place four methods, each isolating one lever against a matched control rather than against B0 alone: M1 for output structure, M2 for external knowledge, M3 for reasoning structure and M4 for reasoning robustness, all detailed in the next section. Closed-book and open-book are an \emph{information switch} layered on top rather than methods of their own.

\subsubsection{Data Construction and Leakage Control}
Everything but model inference runs locally and offline, with no web access at any stage and no local GPU. Parsing the 4963 valid leaves gives the EC tree and hence the valid children of every node, the candidate source at each level for M1 and M2. For each test enzyme we generate cascaded multiple-choice items for levels one through four, each holding the ground-truth path, distractors and an \texttt{UNSURE} option; a level's options are rebuilt at run time from the node just committed, so the ground truth is on offer only while the walk is still on the gold path. We search the query against the \emph{training database} through five offline evidence channels, namely BLAST, HMMER, ESM-kNN, structure and active site, with each hit recording a neighboring enzyme, its EC number and a similarity score. The setting is strictly inductive: \emph{the retrieval database contains only the training set, and test samples never retrieve one another}; UniProt IDs are anonymized so that memorization cannot earn free points. The four splits are low-homology \texttt{<30\%} (432 items), intermediate \texttt{30-50\%} (560), adversarial \texttt{price} (148 items whose annotations were corrected experimentally) and multi-label \texttt{promiscuous} (209), giving \textbf{1349} items and about 5396 level-wise decisions.

\subsubsection{Difficulty Grading}
Every sample is scored along four orthogonal axes normalized to $[0,1]$ (homology $h^{\text{split}}$, class frequency $h^{\text{tail}}$, branching factor $h^{\text{branch}}$ and distractor hardness $h^{\text{distr}}$) and binned by global quintile, from D1 as the easiest to D5 as the hardest, using their unweighted mean $d_i=\tfrac14(h^{\text{split}}_i+h^{\text{tail}}_i+h^{\text{branch}}_i+h^{\text{distr}}_i)$. Distractor hardness can be varied on its own: swapping distractors from distant branches to siblings under the same parent stress-tests fine-grained discrimination \emph{without changing what knowledge is required}. Note that this prior score does \emph{not} account for homology availability in the training database, whereas measured accuracy is driven mainly by homology coverage. This mismatch is itself a finding, and we return to it as the law of homology availability.

\section{Method: One Baseline, Four Levers}

\subsection{B0: Zero-Shot Direct Generation}
B0 reproduces the catastrophic setting of earlier work: the model sees the query, that is the sequence together with the intrinsic descriptors of length, mass, hydropathy, charge and composition we call \emph{derived features}, plus retrieved evidence when open-book, and writes an EC number as free text in a single pass, with no hierarchy, no candidate set and no room to abstain. We parse the EC number from its output, score it with the same hierarchical metrics as every other method, and also report the valid-EC rate.

\subsection{M1: Hierarchical Cascade (Output Structure)}
M1 combines least-to-most decomposition~\cite{zhou2022least} with root-to-leaf navigation of the EC tree. Rather than asking what the EC number is, we ask at each level which of the valid candidates to take, so that the choice at one level constrains the candidates at the next. Every step is a multiple-choice question: the model gives two to four sentences of biochemical reasoning, then picks one \emph{valid candidate} or selects \texttt{UNSURE} to stop at the parent node, and reports a confidence score. A single open generation over 4963 classes thus becomes a closed choice among $7$ candidates at level one and at most $10$ below, invalid EC hallucination drops to zero by construction, although well-formed output does not imply semantic correctness~\cite{park2024grammar}, and partial credit becomes available at every step. Stopping at the deepest reliable ancestor is the decision-theoretic optimum for hierarchical classification with a reject option~\cite{ramaswamy2015convex}, but abstention cuts both ways, as Finding~2 shows.

\subsection{M2: Agentic Search (External Knowledge)}
Closed-book and open-book form an \emph{information switch}, not a method. The evidence comes from a database prebuilt offline under the inductive setting, matching every query against the training database \emph{alone} through five channels: BLAST and HMMER homology, ESM-kNN representation neighbors, Foldseek structure and Folddisco active sites~\cite{kim2026structural}. Each hit returns a neighboring enzyme with its EC number and similarity score. \emph{Passive open-book} pours all hits into the prompt at once, whereas \emph{M2 agentic search} leaves it to the model to decide whether to query, which channel and how far to go, inside a budgeted ReAct loop~\cite{yao2022react}. M2 sits on the cascade backbone of M1 and runs once per level: it forms a hypothesis, judges whether the evidence suffices, and either commits to a valid candidate or names one tool, reads back the observation and updates, with at most $B{=}4$ calls per enzyme. M2 therefore reports \emph{retrieval cost} alongside accuracy, and comparing it with passive open-book over the same cache isolates the net effect of selective retrieval.

\subsection{M3: Mechanistic CoT (Reasoning Structure)}
M3 works on \emph{how} the model thinks, that is, the observable organization of its reasoning rather than its internal computation~\cite{turpin2023language}. It does not walk the EC tree and adds no retrieval of its own, so output structure and external knowledge are both stripped away. The idea is to use the EC number itself as the skeleton of the chain: its digits~\cite{mcdonald2009explorenz} run from reaction class through the bond, group or donor to the acceptor or cofactor and the substrate, a template whose lower levels are class-dependent and hence only heuristic, and M3 makes the model reason along these four axes in order, \emph{specified by hand} rather than composed by the model as in SELF-DISCOVER~\cite{zhou2024self}. Every step is anchored to \emph{intrinsic sequence signals}, offline pattern scans for catalytic and cofactor motifs such as the Rossmann \texttt{GxGxxG}, read off the query sequence with no database lookup and stated in the prompt to be weak priors, not facts. They are one input B0 lacks, so M3 $-$ B0 also credits surfacing them; that can only favor M3, which still loses to closed-book B0 on four of five models. M3$\cdot$closed is therefore a strictly closed-book counterpart to knowledge-graph-anchored chains~\cite{lyu2025knowledge}, while M3$\cdot$open reads the same passive cache for a matched contrast, and both sharpen the PFUA~\cite{fan2026interleaved} result that text-only chains of thought do not help.

\subsection{M4: Self-Consistent CoT (Robustness)}
A single reasoning chain is fragile: it can take a wrong turn early, put too much weight on a weak motif, or misread a noisy hit. M4 applies self-consistency~\cite{wang2022self} as a thin layer that wraps any base reasoner; every M4 run reported here wraps the M1 cascade, sampling $k$ chains and aggregating them by \emph{prefix voting}. Let $v_\ell$ be the majority node committed at level $\ell$ and $f_\ell$ its vote share; descending level by level we accept $v_\ell$ if and only if $f_\ell\ge\tau$, with $\tau=0.5$ by default, and $v_\ell$ extends the previous consensus, otherwise we stop there. The output is the \emph{deepest stable prefix} the chains agree on: agreement commits deeper, disagreement cuts the loss, which directly treats forced and unfounded confidence. M3 changes the content and order of the reasoning while M4 changes the decision rule at test time, so the two are orthogonal. The cost is $k$ times as many calls; the speed of our endpoints kept us to a fixed $k{=}3$.

\section{Evaluation Protocol and Metrics}
We report the four splits separately rather than sweeping one configuration across all of them. The main metric is level-wise \texttt{ori\_acc}: a level counts as correct when the model has committed to it and the node there matches one of the ground-truth prefixes, an any-label match rather than a set-level score,
\begin{equation}
\mathrm{acc}_\ell=\frac1N\sum_{i=1}^{N}\mathbb 1\!\big[\hat n^{(i)}_\ell\ \text{committed}\ \wedge\ \hat n^{(i)}_\ell\in G_\ell(Y_i)\big],
\end{equation}
where $G_\ell(Y)$ is the set of level-$\ell$ ground-truth prefixes. We also report hierarchical precision, recall and F1, which give partial credit for correctly predicted ancestors instead of treating all leaf errors alike~\cite{kosmopoulos2015evaluation}, together with macro-L4, which weights the 7 top-level classes equally. Four diagnostics are where this benchmark adds value: the \textbf{valid-EC rate} for hallucinated EC numbers, the \textbf{calibration error ECE}~\cite{guo2017calibration} for overconfidence, the \textbf{mean stopping depth} \texttt{avg\_commit} for how willing a model is to answer, the \textbf{selective prediction AURC}~\cite{geifman2019selectivenet} for how well a model knows when it does not know, and \textbf{retrieval cost} for what an M2 search strategy returns per query. This batch uses one seed; error bars are $95\%$ bootstrap confidence intervals from 1000 sample-level resamples.

\section{Experiments}
We evaluate five reasoning models with chain of thought on by default: deepseek-v4-pro, gpt-5.5, qwen3-235b-a22b-thinking, gemini-3.1-pro and glm-5.2. All run the same nine settings of Table~\ref{tab:main} on the full 1349 items with no block-level failure; details are in the appendix.

\subsection{Main Results}
Table~\ref{tab:main} collects the full metrics for the five models under nine settings and is our main result. Two things stand out. Every closed-book setting collapses and every open-book setting lands in a narrow band, so the information switch dominates both the choice of method and the choice of model; Figure~\ref{fig:ceiling}(b) draws that gap per model, and within any one open-book setting the models span under $0.06$ in L4, glm-5.2 under M3 aside. The cascade methods, meanwhile, have a valid-EC rate of $1.0$ by construction, yet their closed-book accuracy is \emph{lower} than B0: removing hallucination is not the same as improving accuracy, and a benchmark that reports only leaf accuracy cannot tell the two apart.

\begin{table*}[t]
\centering
\footnotesize
\setlength{\tabcolsep}{4.2pt}
\begin{tabular}{l ccccc c ccc ccc}
\toprule
\multirow{2}{*}{Setting} & \multicolumn{6}{c}{L4: full EC accuracy (per model \& mean)} & \multicolumn{6}{c}{Five-model mean} \\
\cmidrule(lr){2-7}\cmidrule(lr){8-13}
 & DS & GPT & Qwen & Gem & GLM & Mean & L1 & L2 & L3 & valid & H-F1 & ECE \\
\midrule
\multicolumn{13}{l}{\textit{Closed-book (sequence and derived features only)}}\\
B0$\cdot$closed        & 0.159 & 0.036 & 0.010 & 0.084 & 0.013 & 0.060 & 0.352 & 0.173 & 0.122 & 0.803 & 0.164 & \\
M1$\cdot$closed        & 0.047 & 0.059 & 0.002 & 0.123 & 0.004 & 0.047 & 0.290 & 0.142 & 0.092 & 1.000 & 0.146 & 0.608 \\
M3$\cdot$closed        & 0.070 & 0.033 & 0.006 & 0.087 & 0.005 & 0.040 & 0.254 & 0.114 & 0.076 & 0.585 & 0.112 & \\
M4$\cdot$closed        & 0.044 & 0.033 & 0.002 & 0.132 & 0.003 & 0.043 & 0.269 & 0.125 & 0.083 & 1.000 & 0.143 & 0.392 \\
\midrule
\multicolumn{13}{l}{\textit{Open-book (with retrieved evidence)}}\\
B0$\cdot$open        & 0.731 & 0.722 & 0.718 & 0.723 & 0.705 & 0.720 & 0.926 & 0.889 & 0.847 & 0.982 & 0.797 & \\
M2$\cdot$active retrieval    & 0.712 & \textbf{0.747} & 0.712 & 0.709 & \textbf{0.761} & \textbf{0.728} & 0.901 & 0.867 & 0.823 & 1.000 & 0.794 & 0.096 \\
Cascade$\cdot$open      & 0.729 & 0.738 & 0.708 & 0.698 & 0.707 & 0.716 & 0.921 & 0.882 & 0.833 & 1.000 & 0.805 & 0.105 \\
M3$\cdot$open        & 0.716 & 0.718 & 0.695 & 0.707 & 0.599 & 0.687 & 0.884 & 0.848 & 0.808 & 0.931 & 0.760 & \\
M4$\cdot$open        & \textbf{0.733} & 0.741 & \textbf{0.725} & \textbf{0.730} & 0.712 & \textbf{0.728} & 0.922 & 0.889 & 0.843 & 1.000 & 0.813 & 0.100 \\
\bottomrule
\end{tabular}
\caption{\textbf{Main results: five models $\times$ nine settings (full set, $n=1349$).} Left block: L4 accuracy per model and their mean; right block: five-model means. DS, GPT, Qwen, Gem, GLM denote deepseek-v4-pro, gpt-5.5, qwen3-235b, gemini-3.1-pro, glm-5.2. ECE is blank for methods that emit no confidence. Best L4 per column in bold.}
\label{tab:main}
\end{table*}

\subsection{Comparison Against the Specialist Ceiling}
\begin{figure*}[t]
\centering
\includegraphics[width=0.86\textwidth]{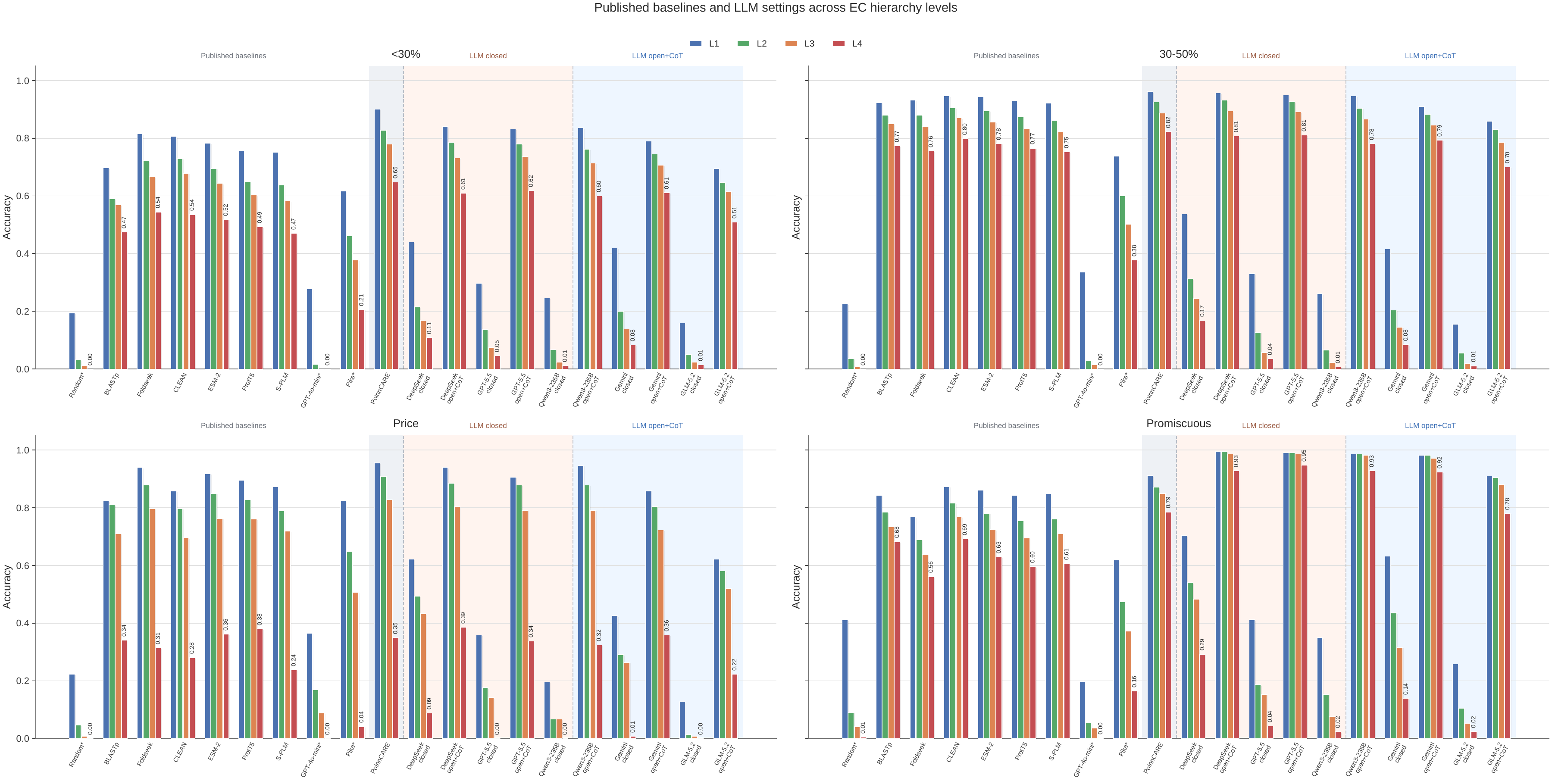}
\caption{\textbf{Level-wise accuracy for published baselines and our LLM settings.} One panel per split; within each method the four bars run from L1 to L4, with L4 labeled on top. Grey: published baselines, in the order random, BLASTp~\cite{altschul1990basic}, Foldseek~\cite{van2024fast}, CLEAN~\cite{yu2023enzyme}, ESM-2~\cite{lin2023evolutionary}, ProtT5~\cite{elnaggar2021prottrans}, S-PLM~\cite{wang2025s}, GPT-4o-mini~\cite{hurst2024gpt}, Pika~\cite{carrami2024pqa} and PoinnCARE~\cite{xie2026poinncare}; starred values are quoted from CARE~\cite{yang2024care}. Orange: closed-book B0. Blue: open-book M3.}
\label{fig:baselines}
\end{figure*}

\begin{figure}[tb]
\centering
\includegraphics[width=0.93\columnwidth]{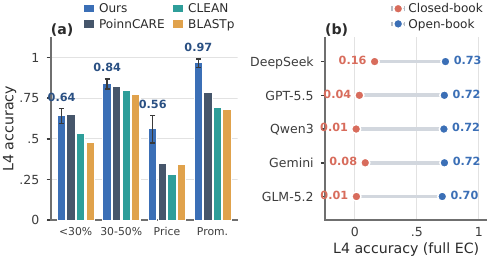}
\caption{\textbf{The specialist ceiling and the closed-to-open jump.} (a) Best open-book LLM against published specialists, per split; \texttt{promiscuous} is not scored alike on the two sides. (b) Closed-book and open-book L4 per model. Bars are $95\%$ bootstrap CIs.}
\label{fig:ceiling}
\end{figure}

Figure~\ref{fig:baselines} places the PoinnCARE baselines next to closed-book B0 and open-book M3 for our five models, on the same splits and inductive setting, the training set being either side's only knowledge. Three patterns sit side by side: specialists decline smoothly, closed-book LLMs collapse after the first level, and open-book reasoning lifts every level back into the specialist band. Reading our best open-book LLM against that ceiling in Figure~\ref{fig:ceiling}(a), we \emph{overtake} every specialist baseline on the adversarial \texttt{price} split, match or slightly exceed PoinnCARE on \texttt{30-50\%}, and fall behind only on \texttt{<30\%}, where our best is $0.64$ against $0.648$, a gap well inside our interval width and one PoinnCARE closes with structure and active sites. We claim nothing on \texttt{promiscuous}, where the two sides are not scored alike: the published numbers score the full label set while we credit any label, which lifts BLASTp from $0.682$ to $0.957$ in our cache, above every LLM setting. We therefore \emph{do not chase state of the art}: with no training, one transferable set of prompts and retrieval strategies takes a general LLM from near zero closed-book to specialist-level accuracy on evidence-rich and adversarial splits.

\subsection{Analysis and Findings}

\subsubsection{Finding 1: Knowledge Is Decisive, and It Repairs More Than Accuracy}
Opening the book multiplies L4 by roughly $5$ to $70$ depending on the model, and the closed-book and open-book $95\%$ intervals are disjoint for every one of them, whereas the five open-book intervals overlap heavily (Figure~\ref{fig:ceiling}(b)). The mechanism is not subtle: the retrieved evidence carries the EC numbers of homologous enzymes, so it puts the answer in the context. On strong models, then, \emph{the EC collapse is caused by knowledge and evidence availability rather than by the task format}, which agrees with the observation that reasoning ability is largely fixed during pretraining~\cite{NEURIPS2025_537d5aa7}.

Evidence repairs more than accuracy, and a leaf-accuracy number hides the rest. For deepseek's M4, opening the book raises consensus depth from $0.76$ to $3.59$ and winning-vote share from $0.772$ to $0.979$, and across the five models the cascade's calibration error falls from $0.42$--$0.72$ closed-book to $0.07$--$0.13$ open-book. Self-consistency pays off only when the base chains are confident enough to agree: closed-book, confused chains vote apart and the method abstains early; open-book, evidence brings them into agreement and the vote carries to the leaf.

\subsubsection{Finding 2: Format Repair Depends on Abstention}
\begin{figure}[tb]
\centering
\includegraphics[width=0.90\columnwidth]{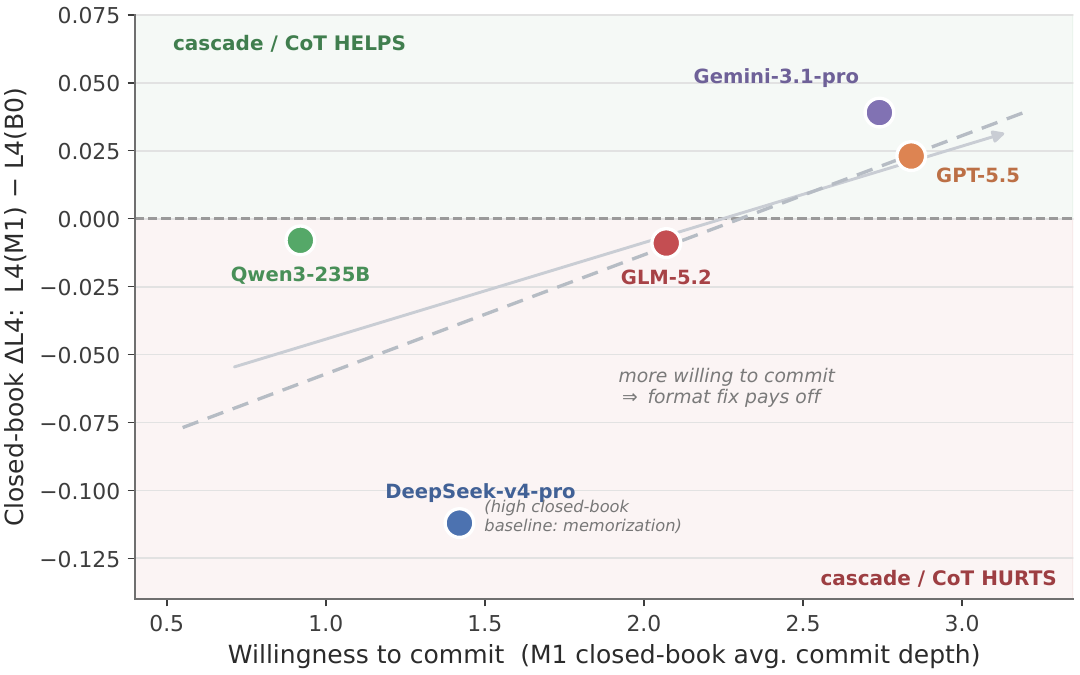}
\caption{\textbf{The closed-book abstention spectrum.} Mean stopping depth of closed-book M1 (larger means more willing to commit) against $\mathrm{L4}(\text{M1})-\mathrm{L4}(\text{B0})$. Deepseek sits below the trend because its closed-book baseline is unusually high, which we suspect reflects memorization.}
\label{fig:abstain}
\end{figure}

Whether repairing the output and reasoning structure, that is closed-book M1, M3 and M4, gains or loses against closed-book B0 is \emph{not universal}; it is decided by how willing a model is to commit. Figure~\ref{fig:abstain} arranges the models along the abstaining-to-insisting spectrum by the mean stopping depth of closed-book M1, and the sign of the net change $\Delta\mathrm{L4}=\mathrm{L4}(\text{M1 closed})-\mathrm{L4}(\text{B0 closed})$ flips along it. For qwen and deepseek, which abstain heavily, \texttt{UNSURE} makes them fall silent early when evidence is missing, and an unanswered deep level counts as wrong, so closed-book M1, M3 and M4 all land \emph{below} B0, with $\Delta\mathrm{L4}=-0.112$ for deepseek. For gemini, which commits readily, closed-book M1 and M4 rise to $0.12$ to $0.13$ and \emph{overtake} B0 at $0.084$. Two observations point to abstention rather than the format itself. The abstentions are well placed: precision conditional on committing at the fourth level is $0.82$--$0.89$ across the five models for M2 and $0.88$--$0.92$ for open-book M4, against $0.18$--$0.73$ for the same models closed-book. Closed-book M4 amplifies whatever tendency a model already has, so gemini gains while qwen abstains even earlier.

\subsubsection{Finding 3: What the LLM Adds Is Arbitration, Not Knowledge}
If evidence decides the outcome, the obvious question is whether the language model earns its place at all. The aggregate answer is unflattering. Voting the EC numbers of the top BLAST and HMMER neighbors from the same offline cache, with no model in the loop, already reaches an L4 of $0.723$ on the full set, while the best of our nine settings reaches $0.728$. Read as a leaderboard entry, five reasoning models buy half a point.

That reading is an artifact of averaging. Table~\ref{tab:netgain} splits the same two numbers four ways: M2 converts $19$ more \texttt{price} items than neighbor voting and loses $19$ \texttt{promiscuous} ones, so the two cancel and the surviving half-point comes from \texttt{<30\%}. The \texttt{price} gain is not a matter of picking the right channel, since it also clears our \emph{oracle best-channel} retrieval bound of $0.426$, which is granted per-sample access to the deepest correct top-1 hit across all five channels; what the model does there is refuse a confident but superseded neighbor, the one move unavailable to a voting rule. On \texttt{promiscuous} the converse holds: plural evidence makes M2 commit to one branch too early, and its loss already shows at level one ($0.89$ against $0.99$ passive); the one pairing that clears neighbor voting on \emph{every} split, glm-5.2 with M2 at $0.761$, draws $55\%$ of its full-set margin from \texttt{price} alone. The sharpest symptom is that M2 and open-book M4 tie \emph{exactly} at $0.728$ on the full set yet differ by $0.07$ on \texttt{price} in one direction and by $0.06$ on \texttt{promiscuous} in the other. One number scores two systems that fail in opposite ways alike; four splits separate them.

\begin{table}[tb]
\centering
\footnotesize
\setlength{\tabcolsep}{3pt}
\begin{tabular}{lccccc}
\toprule
 & \texttt{<30\%} & \texttt{30--50\%} & \texttt{price} & \texttt{promis.} & full \\
$n$ & 432 & 560 & 148 & 209 & 1349 \\
\midrule
Neighbor vote & 0.618 & 0.816 & 0.372 & \textbf{0.938} & 0.723 \\
M2            & 0.634 & 0.817 & \textbf{0.499} & 0.848 & 0.728 \\
M4$\cdot$open & 0.621 & 0.823 & 0.432 & 0.906 & 0.728 \\
\bottomrule
\end{tabular}
\caption{\textbf{L4 against plain neighbor voting.} Neighbor vote takes the EC numbers of the top BLAST and HMMER hits from the same cache with no model in the loop; the LLM rows are five-model means. All three tie overall but disagree sharply where evidence misleads or multiplies.}
\label{tab:netgain}
\end{table}

\subsubsection{Corollary: The Method Matters Only Where Evidence Is Problematic}
Finding~3 has a corollary that sharpens the familiar claim that reasoning wrappers stop mattering once evidence is available. That claim holds only on the two homology-graded splits, where the spread across the five open-book settings averages $0.05$, the width of the confidence intervals themselves. On the two splits where evidence misleads or multiplies it averages $0.16$, and the ranking there is stable rather than random: M2 is the best open-book setting on \texttt{price} for four of the five models and the worst on \texttt{promiscuous} for four of them. \emph{Selective retrieval and premature commitment are one mechanism seen from two sides}, so the wrapper should be chosen for the expected evidence regime rather than from an aggregate leaderboard.

\begin{figure}[tb]
\centering
\includegraphics[width=0.93\columnwidth]{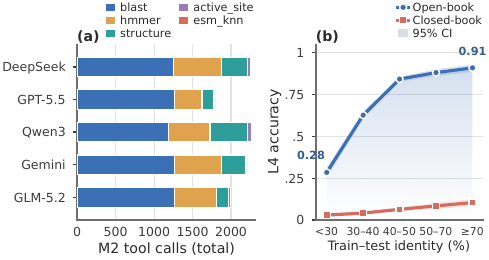}
\caption{\textbf{Where M2 searches, and what homology buys.} (a) M2 tool calls per channel over the full set. (b) L4 against the highest train-test BLAST identity, all models pooled.}
\label{fig:retrieval}
\end{figure}

\begin{figure}[tb]
\centering
\includegraphics[width=0.90\columnwidth]{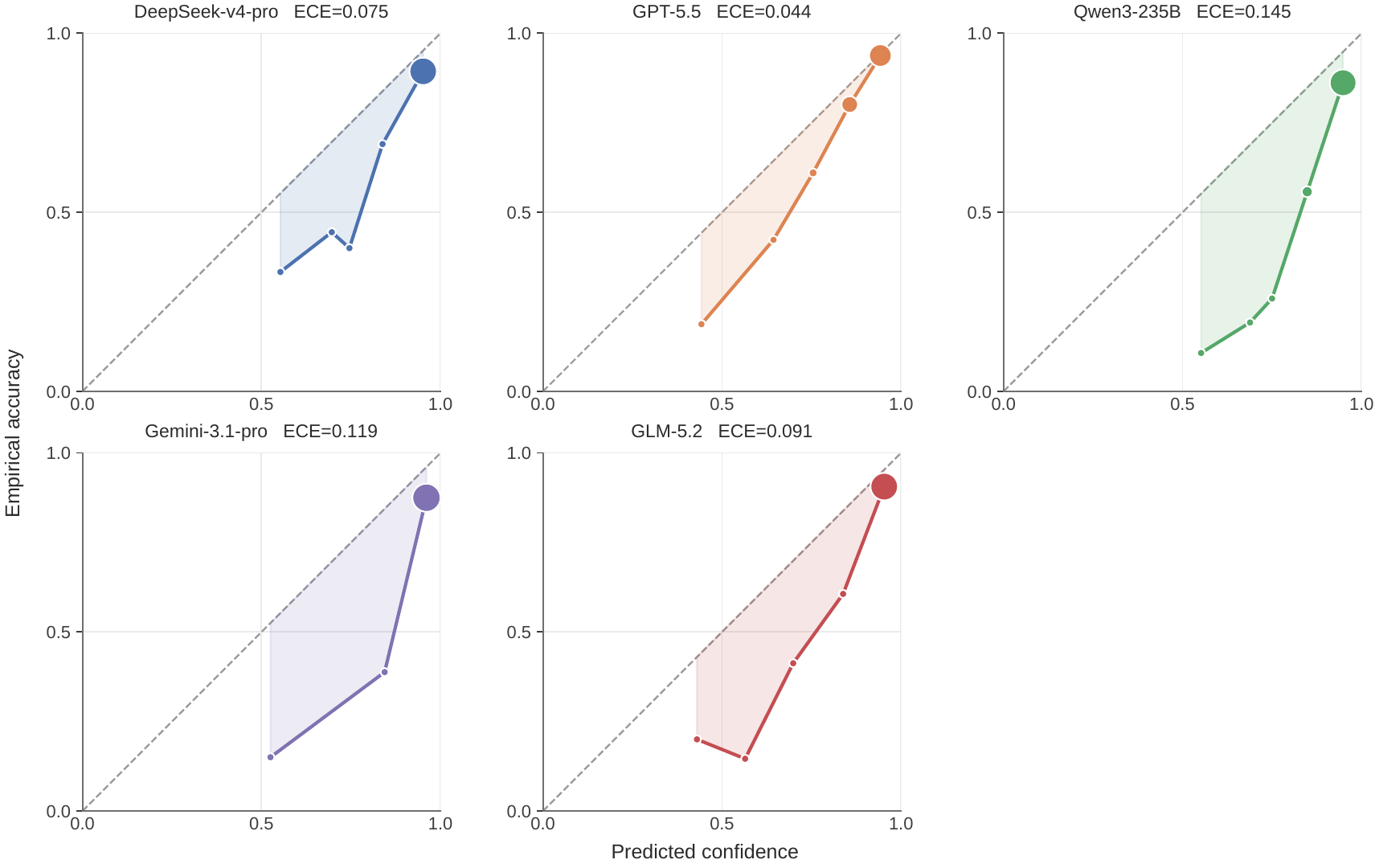}
\caption{\textbf{M2 reliability diagrams.} Self-reported confidence vs. accuracy per bin, one panel per model; the dashed line is perfect calibration and titles give ECE over the bins drawn, within $0.003$ of the all-bin value tabulated in the appendix.}
\label{fig:calib}
\end{figure}

Cost and calibration separate the methods even where accuracy does not. M2 issues only $1.3$ to $1.7$ retrievals per enzyme and is the best-calibrated setting we measured, and its self-reported confidence is usable untouched: the reliability diagrams of Figure~\ref{fig:calib} track the diagonal with no training-time calibration, and answering only the most confident portion raises selective L4 to about $0.85$ to $0.90$. Against that, the five models share one blind spot in Figure~\ref{fig:retrieval}(a): \emph{all of them strongly prefer \texttt{blast}, \texttt{hmmer} and \texttt{structure}, and none ever calls \texttt{esm\_knn}, though it is offered whenever cached}. Such disuse looks more like a \emph{routing failure} across channels than an absence of evidence, so a lightweight router over heterogeneous channels is a concrete handle on low-homology samples.

\subsubsection{Finding 4: The Law of Homology Availability}
Binning all 1349 items by the highest BLAST sequence identity between the test enzyme and the training database, open-book L4 \emph{climbs monotonically} with homology while closed-book stays almost flat, as Figure~\ref{fig:retrieval}(b) shows. The open-book curve spans a factor of three from the lowest bin to the highest, and how clever the model is barely registers on it, which is why the open-book ceiling looks set by evidence rather than by the model. Closed-book accuracy creeps up too, a \emph{faint echo of parametric memory} rather than retrieval, since highly homologous enzymes are better studied and commoner in pretraining corpora.

This curve also settles the mismatch flagged when we defined difficulty. Our prior score never looks at what the retrieval database happens to contain, and the consequence is visible in the appendix: for every model, open-book accuracy at D5 is \emph{higher} than at D4. A difficulty grading blind to evidence availability will mis-rank its own samples, which is a caution for benchmark design well beyond this task. Conversely, the \emph{controlled} homology curve does separate capability from data availability, which is what a diagnostic benchmark adds.

\subsubsection{Failure Modes and Practical Implications}
By split, \texttt{promiscuous} is easiest, being homology-rich, and \texttt{<30\%} hardest; \texttt{price} gains least from open-book access because its experimentally corrected annotations leave neighbors carrying superseded EC numbers. By top-level class the models look alike, with the finely divided EC1 and the recently created EC7 weakest for all five, again reflecting data availability rather than capability. Both patterns point to a retriever that detects conflict rather than a better reasoner, and in the meantime EC prediction is deployable as a \emph{cascade with abstention}.

\section{Conclusion}
We present EC-Reason-Bench, a training-free diagnostic benchmark that decouples the catastrophic failure of general LLMs on enzyme classification into four orthogonal levers. Five reasoning models under nine settings yield four reusable conclusions: knowledge is decisive, with closed-book accuracy near zero and an open-book ceiling set by evidence rather than by the model; the sign of closed-book format repair depends on how readily a model abstains; open-book performance is nearly a function of homology availability; and the net contribution of the model over plain neighbor voting is close to zero in aggregate but large and of opposite sign on the two splits where evidence misleads or multiplies. That last result is the case for diagnostic reporting: adversarial splits such as \texttt{price} discriminate best. Future work targets open models and conflict-aware retrieval.

\bibliography{references}

\clearpage
\input{appendix}

\end{document}

%% file: appendix.tex
% This file is included by main.tex after the bibliography.
\appendix

\section{Extended Related Work and Positioning}
This appendix provides extended related work, a summary of the benchmark design, end-to-end case studies, notes on the evaluation pipeline, complete results, and extended analyses. The main text compresses related work to the lines that bear directly on our design; this section restores the broader context and states our position explicitly.

\paragraph{Specialist EC predictors.} Beyond the homology and contrastive methods cited in the main text, hierarchical supervision appears in DeepEC~\cite{ryu2019deep} and ProteInfer~\cite{sanderson2023proteinfer}, and the recent turn toward structure and multiple modalities includes TopEC~\cite{van2025topec}, ProteinF3S~\cite{yuan2025proteinf3s}, MERA~\cite{wu2026multimodal} and Hyper-Enz~\cite{song2026improving}. HIT-EC~\cite{dumontet2026trustworthy} pairs a four-level hierarchical Transformer with evidential outputs, which is the specialist analogue of coupling our M1 output-structure lever with the M4 robustness lever.

\paragraph{Protein representation benchmarks.} Beginning with TAPE~\cite{rao2019evaluating}, remote-homology splits have been used to show that generalization does not follow from ordinary language ability. Later work folds functional text, three-dimensional structure or homology retrieval into the representation itself, as in ProtST~\cite{xu2023protst}, Prot2Text~\cite{abdine2024prot2text}, ProtCLIP~\cite{zhou2025protclip}, SaProt~\cite{su2024saprot}, Kara~\cite{pmlr-v267-zhang25cz} and Protriever~\cite{weitzman2025protriever}. The multi-task benchmark HeMeNet~\cite{han2025hemenet} flattens EC into 538 binary labels, exactly the kind of flat setting our hierarchical protocol sets out to take apart.

\paragraph{LLM agents for protein annotation.} STELLA~\cite{xiao2026stella} and TIGER~\cite{zhang2026tiger} address function annotation and retrieval through joint sequence and structure encoding and through protein-to-text generation, and the latter documents a failure to generalize under distribution shift.

\paragraph{Knowledge support for LLM reasoning.} In knowledge-intensive tasks, the apparent reasoning ability of an LLM is inseparable from the coverage and organization of the knowledge it can access. Missing or stale facts can force the model to bridge gaps with plausible but unsupported assumptions, and such errors can compound across multi-step inference. Broad, current and coherently structured knowledge is therefore important for constraining intermediate decisions and making final predictions more dependable~\cite{11623696,li2026learning,10884893}.

\paragraph{Inference-time paradigms.} Our levers also relate to tree of thoughts~\cite{yao2023tree}, the on-demand tool calls of Toolformer~\cite{schick2023toolformer} and the self-critique of Self-RAG~\cite{asai2024self}. HiDEC~\cite{im2023hierarchical} recasts hierarchical classification as the generation of a sequence of sub-hierarchies, and zero-shot hierarchical classification without parameter updates~\cite{bhambhoria2023simple} is the closest training-free precedent for M1.

\paragraph{Positioning.} Table~\ref{tab:position} contrasts EC-Reason-Bench with the four most closely related benchmarks and systems along the seven properties that matter for a diagnostic protocol.

\begin{table*}[t]
\centering
\small
\setlength{\tabcolsep}{5pt}
\begin{tabular}{lccccccc}
\toprule
\makecell[l]{Benchmark\\/ method} & \makecell{Hierarchical\\evaluation} & \makecell{Active\\retrieval} & \makecell{Training-\\free} & \makecell{Matched\\ablation} & \makecell{Abstention \&\\calibration} & \makecell{Inductive \&\\leak-free} & \makecell{Multiple\\LLMs} \\
\midrule
CARE~\cite{yang2024care}            & \cmark & \xmark & \cmark & \xmark & \xmark & \cmark & Partial \\
PoinnCARE~\cite{xie2026poinncare}       & \cmark & Partial & \xmark & \xmark & \xmark & \cmark & \xmark \\
PFUA~\cite{fan2026interleaved}            & Partial & \cmark & \cmark & Partial & \xmark & Partial & \cmark \\
Pika~\cite{carrami2024pqa}            & \cmark & Embedding & \cmark & \xmark & \xmark & Partial & \xmark \\
\textbf{EC-Reason-Bench (ours)} & \cmark & \cmark & \cmark & \cmark & \cmark & \cmark & \cmark~(5) \\
\bottomrule
\end{tabular}
\caption{Positioning against prior work. Only EC-Reason-Bench combines all seven properties.}
\label{tab:position}
\end{table*}

\section{The Baseline and the Four Levers at a Glance}
Table~\ref{tab:methods} summarizes the single variable each method changes, its default input, its mechanism, its cost and what it isolates. Each method is described in full in the main text; the table is reproduced here as a compact reference for reading the result tables below.

\begin{table*}[t]
\centering
\footnotesize
\setlength{\tabcolsep}{5pt}
\begin{tabular}{llllccl}
\toprule
Method & Lever & Default input & Core mechanism & valid-EC & Extra cost & Isolates \\
\midrule
B0 Zero-shot & Baseline & \makecell[l]{Sequence\\+ features} & \makecell[l]{One-shot free\\generation} & $<1$ & $1\times$ & \makecell[l]{Catastrophic baseline\\\& hallucination} \\
\makecell[l]{M1 Hierarchical\\cascade} & \makecell[l]{Output\\structure} & \makecell[l]{Sequence\\+ features} & \makecell[l]{Level-wise\\constrained choice} & $1.0$ & $\sim$4 steps & \makecell[l]{Gain from\\output structure} \\
M2 Agentic search & \makecell[l]{External\\knowledge} & \makecell[l]{+ retrieved\\evidence} & \makecell[l]{Self-directed\\ReAct retrieval} & $1.0$ & \makecell{Multi-turn\\+ retrieval} & \makecell[l]{Gain from\\active retrieval} \\
\makecell[l]{M3 Mechanistic\\CoT} & \makecell[l]{Reasoning\\structure} & \makecell[l]{+ intrinsic\\motifs} & \makecell[l]{Reasoning over four\\biochemical axes} & $<1$ & $1\times$ & \makecell[l]{Gain from\\reasoning structure} \\
\makecell[l]{M4 Self-consistent\\CoT} & \makecell[l]{Reasoning\\robustness} & \makecell[l]{Wraps M1\\(as reported)} & \makecell[l]{$k$ chains\\+ prefix voting} & \makecell{Follows\\base} & $\sim k\times$ & \makecell[l]{Gain in robustness\\\& calibration} \\
\bottomrule
\end{tabular}
\caption{One baseline and four training-free levers. Each row targets one conceptually distinct lever, but only M1 and M3 are single-variable interventions relative to B0: M2 also inherits the cascade, and M4 also inherits its base reasoner. Each lever is therefore read against a matched control rather than against B0 alone, namely output structure as M1$\cdot$closed $-$ B0$\cdot$closed, passive external knowledge as B0$\cdot$open $-$ B0$\cdot$closed, active retrieval as M2 $-$ cascade$\cdot$open, reasoning structure as M3 $-$ B0 within the same information setting, and robustness as M4 $-$ M1 within the same information setting. The M3 contrast is matched on the information switch but not on the prompt, since M3 alone is shown the intrinsic motif scan; the notes below quantify how little that can buy. The M4 rows reported in this paper all wrap the M1 cascade; M4 over M3 was not run.}
\label{tab:methods}
\end{table*}

\paragraph{Details omitted from the main text.} The offline construction pipeline additionally exports difficulty labels and a decontamination list alongside the items and the evidence cache, all shared by every method, and a small control subset keeps UniProt IDs visible in order to separate memorization from reasoning. The $4963$ leaves are the EC list vendored with the CARE release~\cite{yang2024care}, all complete four-level numbers, $16$ of them IUBMB preliminary \texttt{n} entries such as \texttt{2.1.1.n7}, and $4936$ of them carrying at least one sequence in the training split; we do not re-derive the list, so the tree inherits CARE's snapshot of the nomenclature rather than a fresh ExplorEnz~\cite{mcdonald2009explorenz} download, and the four-level template of M3 is a hand-written heuristic rather than a claim about the official semantics of levels two to four. Each cascade step offers the $7$ top-level classes at level one and at most $10$ candidates below, with about $70\%$ of the distractors drawn from siblings under the current parent and the rest from other subtrees at the same level, plus an \texttt{UNSURE} option; every candidate is a valid EC node, which is what makes valid-EC $=1.0$ structural. Only the offline stages, that is item construction, the evidence cache and scoring, run without a network; model inference itself is issued to remote OpenAI-compatible endpoints, as the notes below record. In M2 the ReAct loop is capped at three turns per level in addition to the global budget of $B{=}4$ tool calls per enzyme, and the tool menu shown at each step lists only the channels that hold a cached record for the query enzyme. For M4, adaptively stopping the sampling can cut the $k\times$ cost sharply at almost no loss in accuracy; endpoint throughput kept this round at a fixed $k{=}3$. ECE is computed over 10 equal-width confidence bins, and a mean stopping depth of $0$ means the model abstains already at the first level.

\paragraph{How a level's candidates are built, and why they cannot leak the answer.} Two candidate sets exist in the pipeline and should not be confused. The offline item generator stores, for every level, a candidate set built from the \emph{gold} parent, so that stored set does contain every ground-truth node at that level; it is what the distractor-hardness axis of the difficulty score is measured on, and it supports a static, teacher-forced-parent variant of the task that we use only for inspection. Every number reported in this paper instead comes from the cascade evaluator, which rebuilds the options of level $\ell$ from the node the model itself committed at level $\ell-1$: about $70\%$ of the distractors are the other children of that node, the rest are nodes at the same level drawn from elsewhere in the tree, and a ground-truth node is offered only when the committed node is its parent, all ground-truth nodes that qualify being offered together for multi-label enzymes. The two claims in the main text are therefore both claims about the cascade, and they are consistent: an earlier choice does restrict what is offered next, and the ground truth is on offer only while the walk is still on the gold path. Once a model leaves that path the correct node is absent from every later option list, an early error cannot be repaired, and the format cannot hand back an answer the model has already lost --- Case Study 3 is an instance, where M1 commits to isomerases at level one and then walks to \texttt{5.1.1.3} without ever being shown \texttt{1.5}. One consequence is worth stating: because the options depend on the committed parent, two models that diverge at level one face different level-two questions, which is intrinsic to a tree walk and is why \texttt{ori\_acc} scores each level against the ground-truth prefix rather than against a shared option list.

\paragraph{What a closed-book prompt contains.} The closed-book input is sequence-intrinsic throughout. It holds the amino-acid sequence itself, truncated at $1100$ residues with a note giving the full length, and one fixed block of descriptors computed from that sequence offline: length in residues and molecular weight in kDa; mean Kyte-Doolittle hydropathy, hydrophobic fraction, longest hydrophobic run and a membrane-propensity flag; the fractions of positive and negative residues and the net charge; and the three most frequent amino acids with the composition entropy in bits and a low-complexity flag. Nothing else is present: no homology hit, no structure hit, no active-site record and no database domain annotation ever reaches a closed-book prompt, which states in place of the evidence block that none is available. The \texttt{pfam} channel exists only as an open-book retrieval channel and was never populated, so no Pfam or InterPro annotation enters this benchmark at all. Where panel~(C) of Figure~2 in the main text writes \emph{domains} among the derived features, it refers to the intrinsic motif patterns described next rather than to any annotation, and the descriptor block is exactly the one listed above, which carries no isoelectric point.

M3 is the one method that additionally receives the output of the intrinsic motif scanner: a fixed list of six regular expressions --- the Rossmann \texttt{GxGxxG}, the Walker-A P-loop, the $\alpha/\beta$-hydrolase elbow \texttt{GxSxG}, the Zn-metallopeptidase \texttt{HExxH}, the aspartic-protease \texttt{D[TS]G} and the thioredoxin \texttt{C[GP][PH]C} --- matched against the query sequence with no database lookup and no homology search, and labeled in the prompt as weak priors rather than facts. This is an input that B0, M1 and M4 do not get, so the M3 $-$ B0 contrast is matched on the information switch but not on the prompt, and it is the one place where our levers are not input-matched. The direction of the confound is reassuring rather than the reverse: the extra input can only help M3, yet closed-book M3 lands \emph{below} closed-book B0 for four of the five models in Table~\ref{tab:permodel}, and Case Study 3 shows a false-positive \texttt{D[TS]G} match dragging M3 out of the oxidoreductases altogether. A prompt-matched control, B0 with the same motif list appended, would isolate reasoning structure more cleanly, and we did not run it.

\paragraph{What open-book access is, and what it is not.} Open-book means one thing only: the prompt may quote hits from a static evidence cache built offline from the \emph{training split alone}. BLAST, HMMER, ESM-kNN, Foldseek and Folddisco were all run locally against that training database before any model was called, so no channel is a live service and none of them reaches the internet; the tool menu offered to M2 lists exactly these cached channels, with no web search, no query to UniProt, BRENDA or KEGG and no literature lookup, and the same holds for passive open-book, which simply pours the cached hits into the prompt. Each hit is therefore a training enzyme with its EC number and a similarity score, test items never retrieve one another, and UniProt IDs are anonymized, so open-book access supplies training-set homology and nothing about the query enzyme itself. This is also what makes the comparison in Figure~\ref{fig:baselines} a fair one rather than a matter of access: the specialist models quoted there are trained on the same CARE training split, so neither side sees anything beyond it, and the gap between closed- and open-book measures whether a general LLM can use that training-set evidence, not whether it can find extra evidence online.

\section{End-to-End Case Studies}
This section traces the complete processing of three real samples to show how evidence and method choices change the decision path. All three are taken from actual gpt-5.5 runs. The layer-wise decisions, confidence scores, and tool calls are recorded in \texttt{out/results/}. The cases below summarize the relevant hits from the offline evidence cache, while lengthy reasoning traces are presented as abridged paraphrases.

\subsection{Case Study 1: One Retrieval Corrects a Confident Closed-Book Error}
The enzyme \texttt{Q8FP91} belongs to the \texttt{<30\%} split, has difficulty D4 and a length of 642 amino acids, and its ground-truth EC number is \texttt{1.1.5.4}. In the closed-book setting, B0 has access to the sequence and a few global descriptors but no external functional evidence. It builds a plausible explanation from superficial Rossmann- and flavin-related signals, but returns the wrong answer, \texttt{EC=1.3.1.34}. It overgeneralizes from a surface-level motif. For the same enzyme, M2 first considers only the closed-book information at the first level of the cascade. Finding insufficient evidence below level two, it issues one BLAST query. The system retrieves the three nearest neighbors, \texttt{P9WJP5}, \texttt{Q9JXD7}, and \texttt{A2C0M6}, from the offline cache. Their identities range from 44\% to 53\%, and all are annotated as \texttt{1.1.5.4}; the cached HMMER evidence agrees. Because the evidence is decisive, M2 stops retrieving and follows the EC hierarchy to \texttt{1.1.5.4} with high confidence at every level. The retrieval cost is one call. This case captures the decisive role of knowledge: a single homology search exposes the correct label.

\subsection{Case Study 2: Biased Evidence Lets a Homolog Mislead the Open-Book Model}
The adversarial enzyme \texttt{WP\_063462990} belongs to the \texttt{price} split, has difficulty D5 and a length of 280 amino acids, and is a member of the short-chain dehydrogenase/reductase family. Its ground-truth label is \texttt{EC 1.1.1.48}. The annotation for this split was corrected experimentally, but homologs in the training database still carry old or incorrect EC numbers. From SDR family signals alone, closed-book B0 predicts \texttt{1.1.1.100}. M2 makes three calls to BLAST, HMMER, and structure search. The hits are scattered across different leaves under \texttt{1.1.1.*}, show no consensus, and none matches the ground truth. M2 ultimately predicts \texttt{1.1.1.311} at level four with a confidence of only 0.47, accurately reflecting the conflicting evidence. Open-book B0 passively receives all evidence and follows the top BLAST neighbor, returning \texttt{1.1.1.175}. Homologs in the \texttt{price} split cause a systematic bias, which explains why this split receives the smallest open-book gain. All three methods reach the correct third level but fail at the fourth. M2's low level-four confidence makes the conflict visible; even when the prediction is wrong, an honest low-confidence answer is more useful than a confident error.

\subsection{Case Study 3: Open-Book Evidence Resolves a Sample That Defeats Every Closed-Book Method}
The enzyme \texttt{G2IQS8} belongs to the \texttt{<30\%} split, has difficulty D5 and a length of 288 amino acids, and is a member of the methylenetetrahydrofolate reductase family. Its ground-truth EC number is \texttt{1.5.1.54}. Two motifs are detected in the sequence, including a false-positive \texttt{D[TS]G} motif that can derail the reasoning. The three closed-book methods traced here all fail and land in different top-level classes. B0 predicts \texttt{1.3.98.1} under oxidoreductases, M1 confidently follows an isomerase branch to \texttt{5.1.1.3}, and M3 is misled by \texttt{D[TS]G}, classifies the enzyme as a hydrolase, and produces no valid leaf. This is a direct example of missing knowledge in the closed-book setting. With access to evidence, M2 first identifies an oxidoreductase and queries BLAST. The only result is a weak \texttt{1.5.1.53} hit at 25.8\% identity, so M2 requests HMMER evidence. Several profile hits converge on \texttt{1.5.1.54}, and their agreement corrects the prediction to the right leaf. Relying on BLAST alone would select the wrong subgroup. Once the evidence is available, passive open-book B0, the open-book cascade, open-book M3, and open-book M4 all converge on \texttt{1.5.1.54}. Open-book M3 also reverses its hydrolase prediction and treats \texttt{D[TS]G} only as a weak prior. This difficult sample makes clear why external knowledge, active retrieval, and reasoning robustness must be evaluated separately.

\section{Notes on Running the Evaluation}
This section records the operational facts about the run that bear on how the numbers should be read.

\paragraph{Endpoints and model identity.} Model inference is issued to remote OpenAI-compatible endpoints. Before any reported condition was run on it, each endpoint was checked against the model it is supposed to serve rather than trusted to describe itself, so no number in this paper rests on a model name echoed back by an endpoint. Concurrency was set separately per endpoint to stay inside its rate limit and affects wall-clock time only, not any prediction.

\paragraph{Coverage and how failures are counted.} Each condition was executed in two passes, first the two homology-graded splits and then the adversarial and multi-label splits, and every metric reported here is recomputed over the merged set of all $1349$ items, including the retrieval and tool-use columns of panel~(a) of Table~\ref{tab:full-diagnostics}, rather than carried over from either pass. Requests are isolated per enzyme, so a sample whose retries are exhausted is recorded as an abstention and scored as one instead of being dropped. Every denominator therefore equals the full split size, and an unanswered level counts as wrong, which is the convention behind the abstention effects of Finding~2. Under this protocol all five models completed all nine conditions on the full set with no block-level failure and no missing item.

\paragraph{Seeds.} As noted in the main text, running five models under nine settings on $1349$ items confined this batch to a single seed, so the intervals we report are sample-level bootstraps and do not capture run-to-run variance from sampling temperature. The $k{=}3$ chains of M4 are the only place where repeated sampling of the same item enters the protocol.

\section{Complete Results}
The main table reports the five-model mean and each model's L4 result. This section provides the underlying experimental tables in the following order: dataset and evidence coverage, all nine conditions on the full dataset, complete layer-wise results for each split, results by difficulty, and retrieval and self-consistency diagnostics. All four splits are evaluated in full, for a total of $n=1349$. A prediction on the multilabel \texttt{promiscuous} split is counted as correct if it matches any CARE label. This is a single-label any-match rule rather than true multi-label prediction: the system emits one EC number per enzyme, so hitting one of several correct leaves suffices and missing the remaining ones is not penalized. It keeps our numbers comparable with the single-label specialist results we quote, but it is a limitation, and set-level precision, recall and F1 over the whole label set would be the stricter protocol. It also implies that the \texttt{promiscuous} losses discussed in the main text cannot be caused by unreported labels; they come from committing to a branch outside the label set, or from abstaining, when the retrieved neighbors disagree. To avoid duplication, the mean result table, main diagnostic table, and figures already shown in the paper are not repeated here.

\subsection{Dataset Size and Evidence Coverage}
Table~\ref{tab:data-coverage} combines two views with the same denominators. Panel (a) gives the size, multilabel status, and difficulty composition of each split, while panel (b) reports coverage by evidence channel in the offline cache. These counts define the denominators for subsequent grouped results and help explain why sequence-homology channels are more stable than structure and active-site channels.

\paragraph{Split labels and homology bins are two different measures.} The four split names are CARE's and follow CARE's own similarity criterion between test and training enzymes, whereas the homology bins in panel~(d) of Table~\ref{tab:full-diagnostics} are computed by us as the maximum local BLAST identity over the cached training hits of a query, taken as $0$ when no hit exists. The two do not coincide. Of the $432$ items in the \texttt{<30\%} split, $67$ have no cached BLAST hit at all and $300$ have a local hit at $\ge 30\%$ identity, with a median of $34\%$; this is why the $<30$ homology bin holds $179$ items rather than $432$. A local high-scoring pair covering part of a sequence can be much more similar than the two sequences are overall, which is exactly why Case Study 1 is a \texttt{<30\%} item whose nearest cached neighbors align at $44$--$53\%$ identity. Split names therefore always refer to CARE's label and homology always refers to our BLAST measure; the two are reported separately and never mixed.

\begin{table*}[t]
\centering
\small
\setlength{\tabcolsep}{4.2pt}
\begin{tabular}{lrrrrrrr}
\multicolumn{8}{c}{\textbf{(a) Dataset Size, Label Type, and Difficulty Composition}} \\
\toprule
split & $n$ & multilabel & D1 & D2 & D3 & D4 & D5 \\
\midrule
\texttt{<30\%}       & 432  & 0   & 14  & 74  & 116 & 127 & 101 \\
\texttt{30--50\%}    & 560  & 0   & 256 & 182 & 83  & 31  & 8 \\
\texttt{price}       & 148  & 3   & 0   & 4   & 28  & 52  & 64 \\
\texttt{promiscuous} & 209  & 209 & 0   & 10  & 43  & 60  & 96 \\
\midrule
Total                 & 1349 & 212 & 270 & 270 & 270 & 270 & 269 \\
\bottomrule
\end{tabular}

\begin{tabular}{lrrrrrr}
\multicolumn{7}{c}{\textbf{(b) Coverage by Channel in the Offline Evidence Cache}} \\
\toprule
split & $n$ & ESM neighbors & BLAST & HMMER & structure & active site \\
\midrule
\texttt{<30\%}       & 432  & 248  & 365  & 376  & 248  & 22 \\
\texttt{30--50\%}    & 560  & 526  & 551  & 556  & 526  & 55 \\
\texttt{price}       & 148  & 127  & 141  & 143  & 127  & 0 \\
\texttt{promiscuous} & 209  & 200  & 209  & 209  & 200  & 22 \\
\midrule
Total                 & 1349 & 1101 & 1266 & 1284 & 1101 & 99 \\
\bottomrule
\end{tabular}
\caption{Composition and evidence coverage of the complete evaluation set. Panel (a) reports split size, label type, and difficulty distribution. Panel (b) counts samples with at least one available record from each evidence channel; channels may overlap.}
\label{tab:data-coverage}
\end{table*}

\subsection{Comparison with Specialist Models and Retrieval Baselines}
Table~\ref{tab:baselines} places two reference baselines in one float. Panel (a) lists specialist models and LLM baselines reported by PoinnCARE on the same four CARE splits. Panel (b) gives direct-retrieval diagnostics from our offline evidence cache without calling an LLM. The \texttt{oracle best-channel} in panel (b) selects, for each sample, the top-1 result across channels with the deepest correct prefix. It is an optimistic upper bound on retrieval rules that commit to one channel's top hit, not an upper bound on the task, which is why a method that can overrule a neighbor may exceed it; it is not included in the method ranking.

\paragraph{The two panels are not scored alike on \texttt{promiscuous}.} The panels share the splits but not the rule for multi-label enzymes. Panel (a) reproduces published numbers, which score a promiscuous enzyme against its full label set, whereas panel (b) and every LLM result in this paper use the any-label match defined above. The size of that difference can be read off a single tool: BLAST top-hit transfer reaches $0.957$ on \texttt{promiscuous} in our cache against the $0.682$ quoted for BLASTp in panel (a), and $0.597$ against $0.475$ on \texttt{<30\%}, whereas the two agree to within $0.032$ on \texttt{30--50\%} and $0.017$ on \texttt{price}, where few or no labels are plural. We therefore read our results against the published specialists only where the rules coincide, and use our own \texttt{BLAST plus HMMER} neighbor vote as the reference on \texttt{promiscuous}. Under that reference no LLM setting reaches plain neighbor voting on that split, which is the loss analyzed in Finding~3 rather than a gain over specialists.

\begin{table*}[t]
\centering
\small
\setlength{\tabcolsep}{4.2pt}
\begin{tabular}{lcccc}
\multicolumn{5}{c}{\textbf{(a) L4 Accuracy of Specialist Models and LLM Baselines}} \\
\toprule
model & \texttt{<30\%} & \texttt{30--50\%} & \texttt{price} & \texttt{promis.} \\
\midrule
PoinnCARE   & \textbf{0.648} & \textbf{0.822} & 0.349 & \textbf{0.785} \\
CLEAN       & 0.535 & 0.798 & 0.280 & 0.691 \\
ESM-2       & 0.518 & 0.781 & \textbf{0.362} & 0.629 \\
Foldseek    & 0.544 & 0.755 & 0.314 & 0.561 \\
BLASTp      & 0.475 & 0.773 & 0.341 & 0.682 \\
GPT-4o-mini & 0.000 & 0.000 & 0.000 & 0.002 \\
Pika        & 0.206 & 0.377 & 0.041 & 0.164 \\
\bottomrule
\end{tabular}

\begin{tabular}{lcccccccc}
\multicolumn{9}{c}{\textbf{(b) Direct-Retrieval Diagnostics: Overall Layer-Wise and Per-Split L4 Accuracy}} \\
\toprule
& \multicolumn{4}{c}{overall layer-wise accuracy} & \multicolumn{4}{c}{per-split L4 accuracy} \\
\cmidrule(lr){2-5}\cmidrule(lr){6-9}
diagnostic & L1 & L2 & L3 & L4 & \texttt{<30\%} & \texttt{30--50\%} & \texttt{price} & \texttt{promis.} \\
\midrule
BLAST only          & 0.913 & 0.886 & 0.843 & 0.713 & 0.597 & 0.805 & 0.358 & 0.957 \\
HMMER only          & 0.923 & 0.892 & 0.852 & 0.725 & 0.623 & 0.809 & 0.378 & 0.957 \\
BLAST plus HMMER    & 0.921 & 0.891 & 0.851 & 0.723 & 0.618 & 0.816 & 0.372 & 0.938 \\
structure only      & 0.711 & 0.680 & 0.654 & 0.527 & 0.352 & 0.691 & 0.122 & 0.737 \\
structure + site    & 0.724 & 0.693 & 0.664 & 0.538 & 0.359 & 0.711 & 0.122 & 0.742 \\
oracle best-channel & 0.933 & 0.907 & 0.867 & 0.758 & 0.662 & 0.843 & 0.426 & 0.962 \\
\bottomrule
\end{tabular}
\caption{Specialist-model reference results and retrieval diagnostics. Panel (a) is taken from the PoinnCARE paper. Panel (b) uses our offline evidence cache and the same \texttt{ori\_acc} scoring rule.}
\label{tab:baselines}
\end{table*}

\subsection{Complete Nine-Condition Results for Five Models}
Panels (a) and (b) of Table~\ref{tab:permodel} provide all nine conditions and nine overall metrics for every model, which are the source values for the means in the main table. ECE is computed only for methods that report confidence; other entries are left blank. A lower stopping depth indicates earlier abstention. The full-EC precision ranges quoted in Finding~2 of the main text are read from panel (a) of Table~\ref{tab:full-diagnostics} for M2 and from panel (b) for closed- and open-book M4.

Panel~(a) already exposes the strongest abstention contrast: qwen stops at mean depths of $0.92$ and $0.45$ for closed-book M1 and M4, substantially earlier than deepseek and gpt-5.5.

Panel~(b) completes that contrast: gemini stops much later, at mean depths of $2.74$ and $1.92$. One entry is worth noting on its own, because it is invisible in the main table: glm-5.2 reaches the highest M2 accuracy of the five models while having the lowest closed-book valid-EC rate of the five, so a model's fluency at emitting well-formed EC numbers unaided says nothing about how well it will use evidence.

\subsection{Complete Layer-Wise Results by Split}
Tables~\ref{tab:persplit-ds}--\ref{tab:persplit-glm52} give one split-wise panel for each LLM. The four splits are paired within each panel, preserving L1--L4 for all nine conditions in 18 data rows. Every value is exported directly from \texttt{out/results/summary\_*.json}. To keep the final model table from occupying a page by itself, the glm-5.2 panel and the subsequent difficulty panel share one large table. These are the source values for Table~\ref{tab:netgain-full} and for the per-split argument of Finding~3 in the main text.

\subsection{Complete Nine-Condition Results by Difficulty}
Panel (b) of Table~\ref{tab:perdiff} compresses two previously separate tables into paired condition groups, while retaining L4 accuracy for all five models, nine conditions, and difficulty levels D1--D5. Panel (a) contains the complete split-wise results for the final model. Difficulty bins are defined by prior rules and do not account for the availability of homologs in the training database, so the lowest value often occurs at D4 rather than D5. Exact values after regrouping by homology are given in Table~\ref{tab:homology-bins}.

\subsection{Active Retrieval and Self-Consistency Diagnostics}
Table~\ref{tab:full-diagnostics} combines the per-model values underlying the main diagnostic table. Panel (a) aligns M2 cost, submission quality, and tool use. Panel (b) compares consensus depth, winning-vote share, and submission quality for closed- and open-book M4. Panel (c) reports bootstrap intervals, and panel (d) gives exact values for homology bins.

\section{Extended Analysis}
This section provides four supporting analyses that are useful but more detailed than the main discussion: the net contribution of the LLM over plain retrieval, selective prediction and confidence calibration, failure modes by top-level EC class, and statistical significance.

\subsection{Net Contribution over Plain Neighbor Voting}
Table~\ref{tab:netgain-full} expands the main-text comparison to every model. The \emph{neighbor vote} reference is the \texttt{BLAST plus HMMER} row of panel (b) of Table~\ref{tab:baselines}: it takes the EC numbers carried by the top hits of the two sequence-homology channels in the same offline cache and applies the same \texttt{ori\_acc} scoring rule, with no LLM in the loop. The LLM columns are read from the per-split tables collected at the end of the appendix.

Three things follow from the table. First, the full-set column is uninformative: neighbor voting, M2 and open-book M4 fall within $0.005$ of one another, so a leaderboard built on this column would rank a pure retrieval script alongside five reasoning models. Second, that agreement is a cancellation rather than an equivalence. Relative to neighbor voting, M2 wins $0.127$ on the $148$ \texttt{price} items and loses $0.090$ on the $209$ \texttt{promiscuous} items, which is $19$ items converted against $19$ items given back. Third, the direction of the trade is a property of the method and not of the model: M2 beats open-book M4 on \texttt{price} for four of the five models and loses to it on \texttt{promiscuous} for four of the five, and M2's five-model mean on \texttt{price} exceeds even the $0.426$ of the \texttt{oracle best-channel} rule in Table~\ref{tab:baselines}, which bounds any rule that commits to one channel's top hit, so the gain cannot be attributed to channel selection.

The same asymmetry governs how much the choice of method matters at all. Averaging, over the five models, the range of L4 accuracy spanned by the five open-book settings gives $0.051$ on \texttt{<30\%} and $0.052$ on \texttt{30--50\%}, against $0.177$ on \texttt{price} and $0.137$ on \texttt{promiscuous}. Method selection is worth roughly three times as much on the two splits where the retrieved evidence is misleading or plural as on the two graded by homology alone.

\subsection{Selective Prediction and Confidence Calibration}
\begin{figure}[tbp]
\centering
\includegraphics[width=0.99\columnwidth]{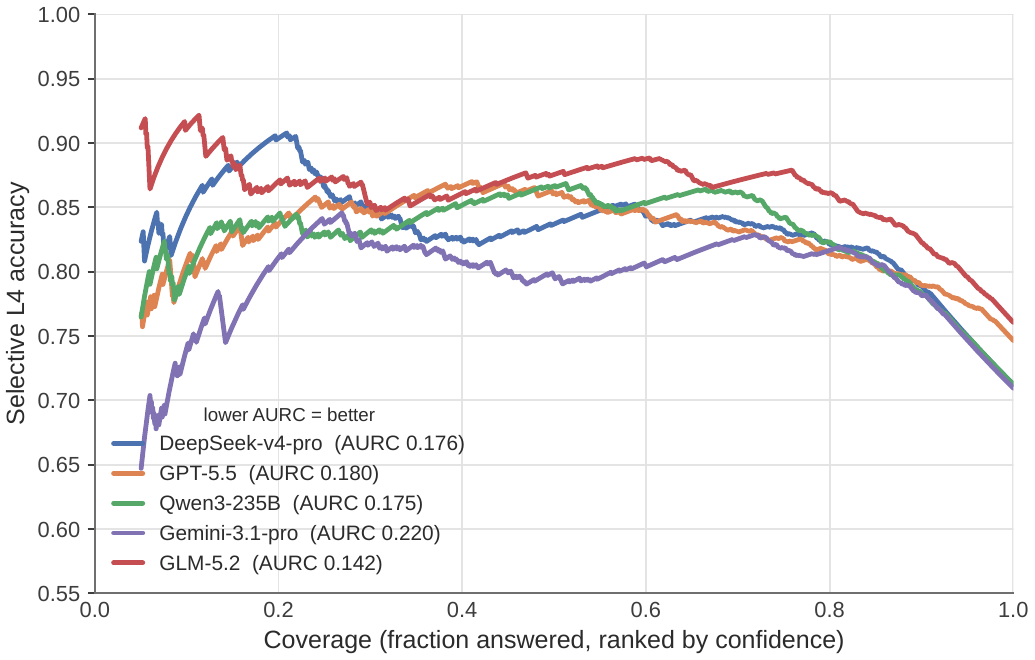}
\caption{Selective prediction with M2 risk-coverage curves. Coverage is the fraction of samples answered by the system, and the vertical axis gives L4 accuracy on those samples. Answering only high-confidence samples raises selective accuracy to about $0.9$. The legend reports AURC, computed on the complementary risk curve as the integral of $1-\text{accuracy}$ over coverage, so a lower AURC is better even though the curve that is drawn is accuracy and higher is better there.}
\label{fig:riskcov}
\end{figure}

The main text includes M2 reliability diagrams because they provide a distinct calibration diagnostic. Here we add the curve most useful for deployment, the risk-coverage curve of Figure~\ref{fig:riskcov}, whose legend reports the area under the complementary risk curve (AURC) per model, lower being better. Confidence is a useful abstention signal for all five models, and the ordering is informative. Gemini ranks last, separating confident from uncertain predictions least effectively, which agrees with its tendency to answer too readily and with its higher ECE in the main-text reliability diagrams. Glm-5.2 ranks first despite having the lowest closed-book valid-EC rate of the five, one more sign that fluency without evidence and calibration under evidence are unrelated properties.

Confidence here is self-reported and has never been calibrated through training. That it is already usable in this raw form is what makes the cascade-with-abstention deployment sketched in the main text practical without any additional trained component, and it suggests temperature scaling or a lightweight calibration head as a cheap further improvement.

\subsection{Failure Modes by Top-Level EC Class}
\begin{figure*}[t]
\centering
\includegraphics[width=0.82\textwidth]{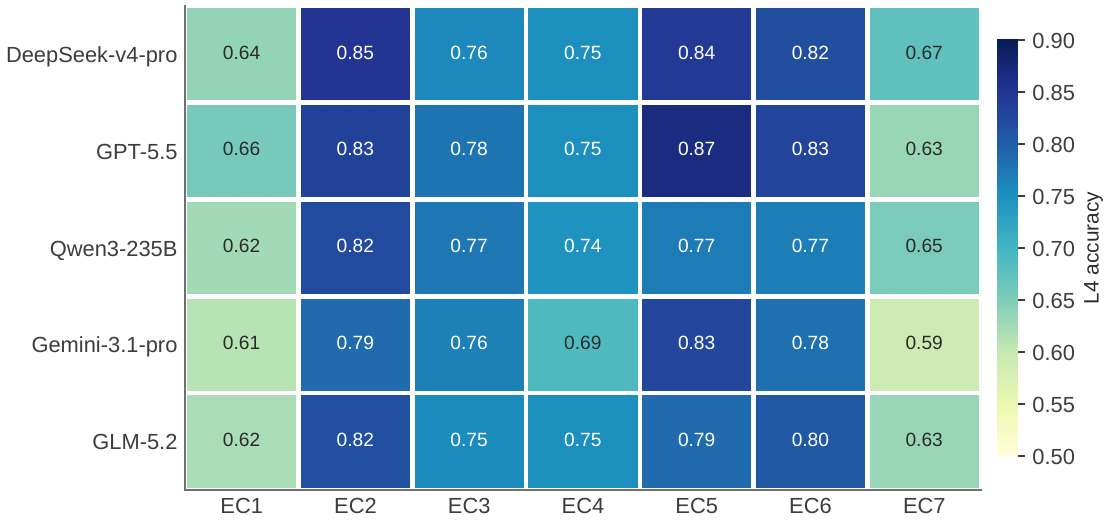}
\caption{L4 accuracy for the open-book cascade across seven top-level EC classes. Rows are models and columns are classes; darker cells indicate higher accuracy. Comparing a column across all five models separates consistently difficult classes from isolated errors. EC1 and EC7 are weak for every model.}
\label{fig:perclass}
\end{figure*}

Figure~\ref{fig:perclass} breaks L4 accuracy for the open-book cascade down over the seven top-level EC classes, and the pattern is highly consistent across all five models: EC2 transferases and EC5 isomerases are strongest, EC1 oxidoreductases and EC7 translocases weakest. The two weak classes fail for different reasons. EC1 is difficult at fine granularity, having the most leaves and many fourth-level distinctions under the same third-level class based on cofactors or substrates, so resolving the final level stays hard even when a close homolog is retrieved. EC7 is instead limited by data: established only in 2018, it is the smallest and youngest top-level class, with sparse coverage in both retrieval databases and pretraining corpora. This again points to data availability rather than model capacity. That the same two classes are weak for every model also shows why macro and micro results should be reported together, since head classes can hide long-tail failures in a micro average.

\subsection{Statistical Significance}
All point estimates carry $95\%$ bootstrap confidence intervals from 1000 sample-level resamples; panel (c) of Table~\ref{tab:full-diagnostics} gives them for closed-book B0, open-book B0 and the best condition of each model, and panel (d) gives the exact values behind the homology curve of the main text. One claim rests on these intervals in the strict sense: the closed- and open-book intervals are disjoint for every model, so the gain from knowledge injection is not sampling noise.

The converse reading is weaker, and we state it as such. Overlapping marginal intervals do not establish equivalence, we did not prespecify an equivalence margin, and we did not run paired between-model tests, so the statement that the open-book ceiling is set by evidence rather than by capability should be read as a description of point estimates that sit close together, not as a demonstrated null: the open-book point estimates lie within $0.06$ of one another in every setting except M3, where glm-5.2 falls to $0.599$ against $0.695$--$0.718$ for the others. A paired bootstrap over per-sample differences would be the right instrument, and its absence is a limitation of the analysis reported here.\newpage
For the same reason we do not claim a tie with the specialists: the best LLM point estimate on \texttt{price} is $0.5608$ from glm-5.2 with M2, and on \texttt{<30\%} it is about $0.64$ against $0.648$ for PoinnCARE, a gap smaller than the width of our own interval, but PoinnCARE's figure is quoted from its paper without an uncertainty estimate and the two numbers were never resampled together.

One further limitation should be stated plainly. The per-split contrasts of Finding~3 are not covered by these intervals, because \texttt{price} holds only $148$ items: the $0.127$ gap to neighbor voting amounts to about $19$ samples, and the matching \texttt{promiscuous} loss to about the same number. We therefore read the direction and the sign consistency across the five models as the evidence, not the magnitude of either gap on its own; the five runs share items and evidence, so they are five correlated observations of one protocol rather than five independent trials.

\FloatBarrier

% Tables 7 onward are collected at the end so explanatory text and figures stay contiguous.

\begin{table*}[p]
\centering
\small
\setlength{\tabcolsep}{4.2pt}
\begin{tabular}{lccccccccc}
\multicolumn{10}{c}{\textbf{(a) deepseek-v4-pro, gpt-5.5, and qwen3-235b}}\\
\toprule
condition & L1 & L2 & L3 & L4 & valid & H-F1 & macro-L4 & ECE & stop depth \\
\midrule
\multicolumn{10}{l}{\textit{deepseek-v4-pro}}\\
B0$\cdot$closed      & 0.541 & 0.337 & 0.278 & 0.159 & 0.904 & 0.304 & 0.162 &       & 3.93 \\
B0$\cdot$open        & 0.932 & 0.900 & 0.857 & 0.731 & 0.993 & 0.806 & 0.768 &       & 3.99 \\
M1$\cdot$closed      & 0.328 & 0.165 & 0.105 & 0.047 & 1.000 & 0.170 & 0.043 & 0.424 & 1.42 \\
M2$\cdot$active      & 0.872 & 0.846 & 0.806 & 0.712 & 1.000 & 0.780 & 0.725 & 0.076 & 3.46 \\
cascade$\cdot$open   & 0.915 & 0.882 & 0.838 & 0.729 & 1.000 & 0.806 & 0.760 & 0.105 & 3.62 \\
M3$\cdot$closed      & 0.334 & 0.168 & 0.124 & 0.070 & 0.689 & 0.161 & 0.062 &       & 3.06 \\
M3$\cdot$open        & 0.924 & 0.889 & 0.847 & 0.716 & 0.974 & 0.795 & 0.754 &       & 3.92 \\
M4$\cdot$closed      & 0.316 & 0.147 & 0.098 & 0.044 & 1.000 & 0.173 & 0.034 & 0.112 & 0.83 \\
M4$\cdot$open        & 0.916 & 0.884 & 0.838 & 0.733 & 1.000 & 0.811 & 0.758 & 0.103 & 3.58 \\
\midrule
\multicolumn{10}{l}{\textit{gpt-5.5}}\\
B0$\cdot$closed      & 0.335 & 0.145 & 0.087 & 0.036 & 0.871 & 0.141 & 0.036 &       & 3.79 \\
B0$\cdot$open        & 0.931 & 0.897 & 0.853 & 0.722 & 0.993 & 0.802 & 0.762 &       & 3.98 \\
M1$\cdot$closed      & 0.345 & 0.166 & 0.102 & 0.059 & 1.000 & 0.169 & 0.051 & 0.619 & 2.84 \\
M2$\cdot$active      & 0.918 & 0.886 & 0.844 & 0.747 & 1.000 & 0.813 & 0.764 & 0.044 & 3.73 \\
cascade$\cdot$open   & 0.925 & 0.892 & 0.850 & 0.738 & 1.000 & 0.817 & 0.763 & 0.068 & 3.70 \\
M3$\cdot$closed      & 0.261 & 0.101 & 0.064 & 0.033 & 0.653 & 0.107 & 0.027 &       & 2.85 \\
M3$\cdot$open        & 0.913 & 0.884 & 0.845 & 0.718 & 0.963 & 0.792 & 0.755 &       & 3.86 \\
M4$\cdot$closed      & 0.333 & 0.141 & 0.079 & 0.033 & 1.000 & 0.162 & 0.021 & 0.502 & 1.68 \\
M4$\cdot$open        & 0.922 & 0.891 & 0.850 & 0.741 & 1.000 & 0.817 & 0.767 & 0.127 & 3.66 \\
\midrule
\multicolumn{10}{l}{\textit{qwen3-235b-a22b-thinking}}\\
B0$\cdot$closed      & 0.262 & 0.080 & 0.036 & 0.010 & 0.831 & 0.090 & 0.012 &       & 3.77 \\
B0$\cdot$open        & 0.926 & 0.889 & 0.845 & 0.718 & 0.989 & 0.796 & 0.757 &       & 4.00 \\
M1$\cdot$closed      & 0.121 & 0.043 & 0.013 & 0.002 & 1.000 & 0.049 & 0.001 & 0.619 & 0.92 \\
M2$\cdot$active      & 0.901 & 0.858 & 0.801 & 0.712 & 1.000 & 0.780 & 0.724 & 0.147 & 3.69 \\
cascade$\cdot$open   & 0.921 & 0.873 & 0.815 & 0.708 & 1.000 & 0.792 & 0.737 & 0.127 & 3.66 \\
M3$\cdot$closed      & 0.301 & 0.090 & 0.032 & 0.006 & 0.890 & 0.098 & 0.005 &       & 3.90 \\
M3$\cdot$open        & 0.917 & 0.868 & 0.827 & 0.695 & 0.990 & 0.778 & 0.731 &       & 4.00 \\
M4$\cdot$closed      & 0.081 & 0.025 & 0.009 & 0.002 & 1.000 & 0.036 & 0.001 & 0.510 & 0.45 \\
M4$\cdot$open        & 0.920 & 0.886 & 0.835 & 0.725 & 1.000 & 0.805 & 0.751 & 0.091 & 3.63 \\
\bottomrule
\end{tabular}

\caption{Complete overall results for nine conditions and five models on all $n=1349$ samples. Panel (a) reports deepseek-v4-pro, gpt-5.5, and qwen3-235b. All entries are rounded from the four-decimal pipeline output, so a value ending in $5$ at the fourth decimal can differ by $0.001$ from a figure recomputed from the split-wise tables.}
\label{tab:permodel}
\end{table*}

\begin{table*}[t]
\ContinuedFloat
\centering
\small
\setlength{\tabcolsep}{4.2pt}
\begin{tabular}{lccccccccc}
\multicolumn{10}{c}{\textbf{(b) gemini-3.1-pro and glm-5.2}}\\
\toprule
condition & L1 & L2 & L3 & L4 & valid & H-F1 & macro-L4 & ECE & stop depth \\
\midrule
\multicolumn{10}{l}{\textit{gemini-3.1-pro}}\\
B0$\cdot$closed      & 0.451 & 0.248 & 0.182 & 0.084 & 0.870 & 0.223 & 0.092 &       & 3.77 \\
B0$\cdot$open        & 0.932 & 0.889 & 0.850 & 0.723 & 0.976 & 0.800 & 0.757 &       & 3.96 \\
M1$\cdot$closed      & 0.410 & 0.255 & 0.205 & 0.123 & 1.000 & 0.240 & 0.119 & 0.655 & 2.74 \\
M2$\cdot$active      & 0.881 & 0.853 & 0.809 & 0.709 & 1.000 & 0.781 & 0.720 & 0.121 & 3.57 \\
cascade$\cdot$open   & 0.920 & 0.876 & 0.827 & 0.698 & 1.000 & 0.801 & 0.721 & 0.122 & 3.62 \\
M3$\cdot$closed      & 0.288 & 0.182 & 0.143 & 0.087 & 0.520 & 0.163 & 0.088 &       & 2.29 \\
M3$\cdot$open        & 0.876 & 0.845 & 0.806 & 0.707 & 0.917 & 0.761 & 0.737 &       & 3.68 \\
M4$\cdot$closed      & 0.409 & 0.258 & 0.208 & 0.132 & 1.000 & 0.255 & 0.130 & 0.400 & 1.92 \\
M4$\cdot$open        & 0.927 & 0.895 & 0.846 & 0.730 & 1.000 & 0.820 & 0.757 & 0.091 & 3.60 \\
\midrule
\multicolumn{10}{l}{\textit{glm-5.2}}\\
B0$\cdot$closed      & 0.170 & 0.057 & 0.025 & 0.013 & 0.537 & 0.061 & 0.015 &       & 2.35 \\
B0$\cdot$open        & 0.907 & 0.870 & 0.828 & 0.705 & 0.960 & 0.780 & 0.748 &       & 3.86 \\
M1$\cdot$closed      & 0.246 & 0.081 & 0.036 & 0.004 & 1.000 & 0.102 & 0.002 & 0.722 & 2.07 \\
M2$\cdot$active      & 0.931 & 0.894 & 0.853 & 0.761 & 1.000 & 0.817 & 0.786 & 0.092 & 3.80 \\
cascade$\cdot$open   & 0.923 & 0.887 & 0.834 & 0.707 & 1.000 & 0.808 & 0.738 & 0.102 & 3.61 \\
M3$\cdot$closed      & 0.084 & 0.030 & 0.016 & 0.005 & 0.174 & 0.030 & 0.008 &       & 0.97 \\
M3$\cdot$open        & 0.788 & 0.755 & 0.717 & 0.599 & 0.813 & 0.672 & 0.639 &       & 3.30 \\
M4$\cdot$closed      & 0.207 & 0.052 & 0.020 & 0.003 & 1.000 & 0.087 & 0.003 & 0.434 & 0.83 \\
M4$\cdot$open        & 0.924 & 0.891 & 0.844 & 0.712 & 1.000 & 0.811 & 0.744 & 0.090 & 3.59 \\
\bottomrule
\end{tabular}
\caption{Complete overall results for nine conditions and five models (continued). Panel (b) reports gemini-3.1-pro and glm-5.2. Both panels give L1--L4, valid-EC rate, H-F1, macro-L4, ECE, and mean stopping depth. ECE uses ten equal-width confidence bins over all samples that report confidence; the reliability diagrams in the main text retain only bins with at least $15$ samples, lowering the displayed value by up to $0.003$.}
\end{table*}

\begin{table*}[t]
\centering
\small
\setlength{\tabcolsep}{4.2pt}
\begin{tabular}{lcccccccc}
\toprule
& \multicolumn{4}{c}{\texttt{<30\%} ($n=432$)} & \multicolumn{4}{c}{\texttt{30--50\%} ($n=560$)} \\
\cmidrule(lr){2-5}\cmidrule(lr){6-9}
condition & L1 & L2 & L3 & L4 & L1 & L2 & L3 & L4 \\
\midrule
B0$\cdot$closed & 0.4398 & 0.2153 & 0.1690 & 0.1088 & 0.5375 & 0.3125 & 0.2446 & 0.1679 \\
B0$\cdot$open & 0.8472 & 0.7963 & 0.7500 & 0.6296 & 0.9625 & 0.9393 & 0.9018 & 0.8161 \\
M1$\cdot$closed & 0.2639 & 0.1134 & 0.0625 & 0.0231 & 0.3196 & 0.1482 & 0.0964 & 0.0571 \\
M2$\cdot$active & 0.8426 & 0.8009 & 0.7593 & 0.6343 & 0.9518 & 0.9321 & 0.8929 & 0.8304 \\
cascade$\cdot$open & 0.8380 & 0.7894 & 0.7407 & 0.6273 & 0.9500 & 0.9286 & 0.8875 & 0.8143 \\
M3$\cdot$closed & 0.2685 & 0.1181 & 0.0764 & 0.0532 & 0.3339 & 0.1464 & 0.1089 & 0.0732 \\
M3$\cdot$open & 0.8403 & 0.7847 & 0.7315 & 0.6088 & 0.9571 & 0.9321 & 0.8946 & 0.8071 \\
M4$\cdot$closed & 0.2523 & 0.1042 & 0.0648 & 0.0347 & 0.3214 & 0.1304 & 0.0839 & 0.0500 \\
M4$\cdot$open & 0.8356 & 0.7870 & 0.7361 & 0.6204 & 0.9589 & 0.9411 & 0.8982 & 0.8357 \\
\midrule
& \multicolumn{4}{c}{\texttt{price} ($n=148$)} & \multicolumn{4}{c}{\texttt{promiscuous} ($n=209$)} \\
\cmidrule(lr){2-5}\cmidrule(lr){6-9}
condition & L1 & L2 & L3 & L4 & L1 & L2 & L3 & L4 \\
\midrule
B0$\cdot$closed & 0.6216 & 0.4932 & 0.4324 & 0.0878 & 0.7033 & 0.5407 & 0.4833 & 0.2919 \\
B0$\cdot$open & 0.9730 & 0.9189 & 0.8108 & 0.3851 & 0.9952 & 0.9952 & 0.9904 & 0.9569 \\
M1$\cdot$closed & 0.3919 & 0.2297 & 0.1689 & 0.0270 & 0.4354 & 0.2727 & 0.1722 & 0.0861 \\
M2$\cdot$active & 0.7770 & 0.7432 & 0.6689 & 0.4392 & 0.7847 & 0.7799 & 0.7656 & 0.7512 \\
cascade$\cdot$open & 0.9527 & 0.8986 & 0.8041 & 0.4459 & 0.9569 & 0.9378 & 0.9330 & 0.9139 \\
M3$\cdot$closed & 0.3851 & 0.2162 & 0.1689 & 0.0338 & 0.4306 & 0.2919 & 0.2297 & 0.1244 \\
M3$\cdot$open & 0.9392 & 0.8851 & 0.8041 & 0.3851 & 0.9952 & 0.9952 & 0.9856 & 0.9282 \\
M4$\cdot$closed & 0.4122 & 0.2432 & 0.1757 & 0.0135 & 0.3636 & 0.2153 & 0.1483 & 0.0670 \\
M4$\cdot$open & 0.9527 & 0.9054 & 0.8176 & 0.4595 & 0.9426 & 0.9187 & 0.8995 & 0.8852 \\
\bottomrule
\end{tabular}
\caption{Complete split-wise layer accuracy for deepseek-v4-pro. The four splits are shown in two pairs, with L1--L4 retained for all nine experimental conditions.}
\label{tab:persplit-ds}
\end{table*}

\begin{table*}[t]
\centering
\small
\setlength{\tabcolsep}{4.2pt}
\begin{tabular}{lcccccccc}
\toprule
& \multicolumn{4}{c}{\texttt{<30\%} ($n=432$)} & \multicolumn{4}{c}{\texttt{30--50\%} ($n=560$)} \\
\cmidrule(lr){2-5}\cmidrule(lr){6-9}
condition & L1 & L2 & L3 & L4 & L1 & L2 & L3 & L4 \\
\midrule
B0$\cdot$closed & 0.2963 & 0.1366 & 0.0741 & 0.0463 & 0.3304 & 0.1268 & 0.0571 & 0.0357 \\
B0$\cdot$open & 0.8495 & 0.7963 & 0.7454 & 0.6157 & 0.9625 & 0.9357 & 0.8946 & 0.8125 \\
M1$\cdot$closed & 0.3241 & 0.1528 & 0.0949 & 0.0671 & 0.3321 & 0.1482 & 0.0875 & 0.0589 \\
M2$\cdot$active & 0.8542 & 0.7940 & 0.7431 & 0.6389 & 0.9500 & 0.9286 & 0.8946 & 0.8304 \\
cascade$\cdot$open & 0.8565 & 0.7986 & 0.7500 & 0.6319 & 0.9625 & 0.9411 & 0.9018 & 0.8304 \\
M3$\cdot$closed & 0.2245 & 0.0856 & 0.0532 & 0.0347 & 0.2554 & 0.0875 & 0.0482 & 0.0339 \\
M3$\cdot$open & 0.8310 & 0.7801 & 0.7361 & 0.6181 & 0.9500 & 0.9268 & 0.8911 & 0.8107 \\
M4$\cdot$closed & 0.3079 & 0.1204 & 0.0602 & 0.0301 & 0.3268 & 0.1321 & 0.0679 & 0.0375 \\
M4$\cdot$open & 0.8472 & 0.7986 & 0.7546 & 0.6412 & 0.9607 & 0.9375 & 0.9018 & 0.8321 \\
\midrule
& \multicolumn{4}{c}{\texttt{price} ($n=148$)} & \multicolumn{4}{c}{\texttt{promiscuous} ($n=209$)} \\
\cmidrule(lr){2-5}\cmidrule(lr){6-9}
condition & L1 & L2 & L3 & L4 & L1 & L2 & L3 & L4 \\
\midrule
B0$\cdot$closed & 0.3581 & 0.1757 & 0.1419 & 0.0000 & 0.4115 & 0.1866 & 0.1531 & 0.0431 \\
B0$\cdot$open & 0.9595 & 0.9054 & 0.8176 & 0.3784 & 0.9952 & 0.9952 & 0.9904 & 0.9426 \\
M1$\cdot$closed & 0.3784 & 0.2095 & 0.1486 & 0.0338 & 0.4019 & 0.2105 & 0.1244 & 0.0622 \\
M2$\cdot$active & 0.9595 & 0.9189 & 0.8311 & 0.5135 & 0.9378 & 0.9378 & 0.9282 & 0.9091 \\
cascade$\cdot$open & 0.9459 & 0.9054 & 0.8176 & 0.4459 & 0.9522 & 0.9426 & 0.9378 & 0.9187 \\
M3$\cdot$closed & 0.2770 & 0.1149 & 0.0878 & 0.0000 & 0.3397 & 0.1579 & 0.1100 & 0.0526 \\
M3$\cdot$open & 0.9054 & 0.8784 & 0.7905 & 0.3378 & 0.9904 & 0.9904 & 0.9856 & 0.9474 \\
M4$\cdot$closed & 0.4054 & 0.2095 & 0.1689 & 0.0203 & 0.3493 & 0.1579 & 0.0813 & 0.0383 \\
M4$\cdot$open & 0.9527 & 0.9122 & 0.8176 & 0.4392 & 0.9522 & 0.9426 & 0.9330 & 0.9187 \\
\bottomrule
\end{tabular}
\caption{Complete split-wise layer accuracy for gpt-5.5. The four splits are shown in two pairs, with L1--L4 retained for all nine experimental conditions.}
\label{tab:persplit-g55}
\end{table*}

\begin{table*}[t]
\centering
\small
\setlength{\tabcolsep}{4.2pt}
\begin{tabular}{lcccccccc}
\toprule
& \multicolumn{4}{c}{\texttt{<30\%} ($n=432$)} & \multicolumn{4}{c}{\texttt{30--50\%} ($n=560$)} \\
\cmidrule(lr){2-5}\cmidrule(lr){6-9}
condition & L1 & L2 & L3 & L4 & L1 & L2 & L3 & L4 \\
\midrule
B0$\cdot$closed & 0.2454 & 0.0671 & 0.0231 & 0.0116 & 0.2607 & 0.0661 & 0.0214 & 0.0071 \\
B0$\cdot$open & 0.8519 & 0.7894 & 0.7292 & 0.6181 & 0.9500 & 0.9268 & 0.8964 & 0.8161 \\
M1$\cdot$closed & 0.0949 & 0.0370 & 0.0046 & 0.0000 & 0.1375 & 0.0464 & 0.0143 & 0.0054 \\
M2$\cdot$active & 0.8333 & 0.7731 & 0.7245 & 0.6366 & 0.9429 & 0.9036 & 0.8429 & 0.7821 \\
cascade$\cdot$open & 0.8380 & 0.7708 & 0.7060 & 0.5949 & 0.9571 & 0.9196 & 0.8589 & 0.7875 \\
M3$\cdot$closed & 0.2569 & 0.0625 & 0.0139 & 0.0023 & 0.2875 & 0.0821 & 0.0196 & 0.0054 \\
M3$\cdot$open & 0.8356 & 0.7616 & 0.7130 & 0.5995 & 0.9464 & 0.9036 & 0.8661 & 0.7804 \\
M4$\cdot$closed & 0.0787 & 0.0255 & 0.0093 & 0.0000 & 0.0857 & 0.0268 & 0.0071 & 0.0018 \\
M4$\cdot$open & 0.8310 & 0.7870 & 0.7315 & 0.6134 & 0.9571 & 0.9232 & 0.8750 & 0.8036 \\
\midrule
& \multicolumn{4}{c}{\texttt{price} ($n=148$)} & \multicolumn{4}{c}{\texttt{promiscuous} ($n=209$)} \\
\cmidrule(lr){2-5}\cmidrule(lr){6-9}
condition & L1 & L2 & L3 & L4 & L1 & L2 & L3 & L4 \\
\midrule
B0$\cdot$closed & 0.1959 & 0.0676 & 0.0676 & 0.0000 & 0.3493 & 0.1531 & 0.0766 & 0.0239 \\
B0$\cdot$open & 0.9527 & 0.8986 & 0.8041 & 0.3514 & 0.9952 & 0.9856 & 0.9761 & 0.9234 \\
M1$\cdot$closed & 0.0743 & 0.0203 & 0.0000 & 0.0000 & 0.1627 & 0.0622 & 0.0335 & 0.0000 \\
M2$\cdot$active & 0.9324 & 0.8986 & 0.7973 & 0.4932 & 0.9043 & 0.8804 & 0.8517 & 0.8373 \\
cascade$\cdot$open & 0.9324 & 0.8514 & 0.7703 & 0.4392 & 0.9856 & 0.9713 & 0.9522 & 0.9187 \\
M3$\cdot$closed & 0.2973 & 0.0811 & 0.0473 & 0.0000 & 0.4306 & 0.1770 & 0.0909 & 0.0191 \\
M3$\cdot$open & 0.9459 & 0.8784 & 0.7905 & 0.3243 & 0.9856 & 0.9856 & 0.9809 & 0.9282 \\
M4$\cdot$closed & 0.0541 & 0.0000 & 0.0000 & 0.0000 & 0.0909 & 0.0335 & 0.0191 & 0.0048 \\
M4$\cdot$open & 0.9392 & 0.8919 & 0.7973 & 0.4730 & 0.9904 & 0.9856 & 0.9713 & 0.9234 \\
\bottomrule
\end{tabular}
\caption{Complete split-wise layer accuracy for qwen3-235b-a22b-thinking. The four splits are shown in two pairs, with L1--L4 retained for all nine experimental conditions.}
\label{tab:persplit-q235}
\end{table*}

\begin{table*}[p]
\centering
\small
\setlength{\tabcolsep}{4.2pt}
\begin{tabular}{lcccccccc}
\toprule
& \multicolumn{4}{c}{\texttt{<30\%} ($n=432$)} & \multicolumn{4}{c}{\texttt{30--50\%} ($n=560$)} \\
\cmidrule(lr){2-5}\cmidrule(lr){6-9}
condition & L1 & L2 & L3 & L4 & L1 & L2 & L3 & L4 \\
\midrule
B0$\cdot$closed & 0.4190 & 0.1991 & 0.1389 & 0.0833 & 0.4161 & 0.2036 & 0.1446 & 0.0839 \\
B0$\cdot$open & 0.8565 & 0.7847 & 0.7384 & 0.6181 & 0.9607 & 0.9286 & 0.8982 & 0.8179 \\
M1$\cdot$closed & 0.3773 & 0.2176 & 0.1574 & 0.0972 & 0.3964 & 0.2339 & 0.1786 & 0.1179 \\
M2$\cdot$active & 0.8194 & 0.7685 & 0.7292 & 0.6204 & 0.9393 & 0.9232 & 0.8768 & 0.8125 \\
cascade$\cdot$open & 0.8588 & 0.7963 & 0.7431 & 0.5995 & 0.9518 & 0.9214 & 0.8804 & 0.7964 \\
M3$\cdot$closed & 0.2685 & 0.1597 & 0.1227 & 0.0856 & 0.2661 & 0.1482 & 0.1054 & 0.0786 \\
M3$\cdot$open & 0.7894 & 0.7454 & 0.7060 & 0.6111 & 0.9089 & 0.8821 & 0.8446 & 0.7929 \\
M4$\cdot$closed & 0.3704 & 0.2269 & 0.1713 & 0.1111 & 0.4036 & 0.2339 & 0.1821 & 0.1268 \\
M4$\cdot$open & 0.8588 & 0.8102 & 0.7593 & 0.6343 & 0.9679 & 0.9500 & 0.9054 & 0.8375 \\
\midrule
& \multicolumn{4}{c}{\texttt{price} ($n=148$)} & \multicolumn{4}{c}{\texttt{promiscuous} ($n=209$)} \\
\cmidrule(lr){2-5}\cmidrule(lr){6-9}
condition & L1 & L2 & L3 & L4 & L1 & L2 & L3 & L4 \\
\midrule
B0$\cdot$closed & 0.4257 & 0.2905 & 0.2635 & 0.0068 & 0.6316 & 0.4354 & 0.3158 & 0.1388 \\
B0$\cdot$open & 0.9527 & 0.8919 & 0.8041 & 0.3649 & 0.9952 & 0.9952 & 0.9856 & 0.9378 \\
M1$\cdot$closed & 0.3851 & 0.2770 & 0.2703 & 0.0811 & 0.5311 & 0.3732 & 0.3301 & 0.2201 \\
M2$\cdot$active & 0.9189 & 0.8986 & 0.8108 & 0.4865 & 0.8230 & 0.8038 & 0.7943 & 0.7751 \\
cascade$\cdot$open & 0.9459 & 0.9122 & 0.8041 & 0.4189 & 0.9426 & 0.8947 & 0.8708 & 0.8373 \\
M3$\cdot$closed & 0.3311 & 0.2297 & 0.2162 & 0.0270 & 0.3589 & 0.2823 & 0.2344 & 0.1531 \\
M3$\cdot$open & 0.8581 & 0.8041 & 0.7230 & 0.3581 & 0.9809 & 0.9809 & 0.9713 & 0.9234 \\
M4$\cdot$closed & 0.4189 & 0.2905 & 0.2703 & 0.0946 & 0.4976 & 0.3636 & 0.3110 & 0.2153 \\
M4$\cdot$open & 0.9392 & 0.8986 & 0.7973 & 0.3851 & 0.9474 & 0.9187 & 0.8995 & 0.8852 \\
\bottomrule
\end{tabular}
\caption{Complete split-wise layer accuracy for gemini-3.1-pro. The four splits are shown in two pairs, with L1--L4 retained for all nine experimental conditions.}
\label{tab:persplit-g31}
\end{table*}

\begin{table*}[t]
\centering
\small
\setlength{\tabcolsep}{4.2pt}
\begin{tabular}{lcccccccc}
\multicolumn{9}{c}{\textbf{(a) Complete Split-Wise Layer Accuracy for glm-5.2}} \\
\toprule
& \multicolumn{4}{c}{\texttt{<30\%} ($n=432$)} & \multicolumn{4}{c}{\texttt{30--50\%} ($n=560$)} \\
\cmidrule(lr){2-5}\cmidrule(lr){6-9}
condition & L1 & L2 & L3 & L4 & L1 & L2 & L3 & L4 \\
\midrule
B0$\cdot$closed & 0.1597 & 0.0509 & 0.0231 & 0.0139 & 0.1554 & 0.0554 & 0.0196 & 0.0107 \\
B0$\cdot$open & 0.8148 & 0.7593 & 0.7176 & 0.5972 & 0.9446 & 0.9143 & 0.8768 & 0.7964 \\
M1$\cdot$closed & 0.2060 & 0.0694 & 0.0301 & 0.0046 & 0.2500 & 0.0661 & 0.0232 & 0.0054 \\
M2$\cdot$active & 0.8588 & 0.7894 & 0.7431 & 0.6389 & 0.9589 & 0.9339 & 0.8964 & 0.8304 \\
cascade$\cdot$open & 0.8356 & 0.7824 & 0.7245 & 0.5972 & 0.9607 & 0.9357 & 0.8893 & 0.8054 \\
M3$\cdot$closed & 0.0856 & 0.0324 & 0.0162 & 0.0093 & 0.0786 & 0.0196 & 0.0071 & 0.0018 \\
M3$\cdot$open & 0.6944 & 0.6458 & 0.6157 & 0.5093 & 0.8589 & 0.8304 & 0.7857 & 0.7000 \\
M4$\cdot$closed & 0.1759 & 0.0440 & 0.0185 & 0.0023 & 0.2107 & 0.0375 & 0.0125 & 0.0036 \\
M4$\cdot$open & 0.8356 & 0.7847 & 0.7361 & 0.5949 & 0.9643 & 0.9393 & 0.8911 & 0.8054 \\
\midrule
& \multicolumn{4}{c}{\texttt{price} ($n=148$)} & \multicolumn{4}{c}{\texttt{promiscuous} ($n=209$)} \\
\cmidrule(lr){2-5}\cmidrule(lr){6-9}
condition & L1 & L2 & L3 & L4 & L1 & L2 & L3 & L4 \\
\midrule
B0$\cdot$closed & 0.1284 & 0.0135 & 0.0068 & 0.0000 & 0.2584 & 0.1053 & 0.0526 & 0.0239 \\
B0$\cdot$open & 0.9189 & 0.8649 & 0.7703 & 0.3716 & 0.9856 & 0.9856 & 0.9665 & 0.9187 \\
M1$\cdot$closed & 0.2703 & 0.1081 & 0.0676 & 0.0000 & 0.3014 & 0.1244 & 0.0622 & 0.0000 \\
M2$\cdot$active & 0.9459 & 0.9122 & 0.8243 & 0.5608 & 0.9952 & 0.9904 & 0.9809 & 0.9665 \\
cascade$\cdot$open & 0.9527 & 0.8851 & 0.7905 & 0.3784 & 0.9809 & 0.9713 & 0.9426 & 0.9043 \\
M3$\cdot$closed & 0.0203 & 0.0000 & 0.0000 & 0.0000 & 0.1388 & 0.0718 & 0.0478 & 0.0096 \\
M3$\cdot$open & 0.6216 & 0.5811 & 0.5203 & 0.2230 & 0.9091 & 0.9043 & 0.8804 & 0.7799 \\
M4$\cdot$closed & 0.1689 & 0.0405 & 0.0135 & 0.0000 & 0.2871 & 0.1148 & 0.0478 & 0.0048 \\
M4$\cdot$open & 0.9459 & 0.9122 & 0.8243 & 0.4054 & 0.9809 & 0.9665 & 0.9522 & 0.9187 \\
\bottomrule
\end{tabular}
\par\vspace{2pt}
\setlength{\tabcolsep}{1.7pt}
\begin{tabular}{llccccc@{\hspace{5pt}}lccccc}
\multicolumn{13}{c}{\textbf{(b) L4 Accuracy for All Nine Conditions by Difficulty}} \\
\toprule
\multirow{2}{*}{model} & \multicolumn{6}{c}{condition group I} & \multicolumn{6}{c}{condition group II} \\
\cmidrule(lr){2-7}\cmidrule(lr){8-13}
 & condition & D1 & D2 & D3 & D4 & D5 & condition & D1 & D2 & D3 & D4 & D5 \\
\midrule
\multirow{5}{*}{\texttt{deepseek-v4-pro}} & B0$\cdot$closed & 0.2407 & 0.1667 & 0.1148 & 0.1259 & 0.1487 & B0$\cdot$open & 0.9259 & 0.7444 & 0.7444 & 0.5889 & 0.6506 \\
 & M1$\cdot$closed & 0.0630 & 0.0519 & 0.0370 & 0.0407 & 0.0446 & M2$\cdot$active & 0.9519 & 0.7259 & 0.7259 & 0.5556 & 0.6022 \\
 & cascade$\cdot$open & 0.9259 & 0.7444 & 0.7259 & 0.5889 & 0.6617 & M3$\cdot$closed & 0.1074 & 0.0741 & 0.0630 & 0.0556 & 0.0520 \\
 & M3$\cdot$open & 0.9259 & 0.7444 & 0.7222 & 0.5667 & 0.6208 & M4$\cdot$closed & 0.0481 & 0.0593 & 0.0333 & 0.0407 & 0.0372 \\
 & M4$\cdot$open & 0.9407 & 0.7481 & 0.7259 & 0.5852 & 0.6654 & N/A & \multicolumn{5}{c}{N/A} \\
\midrule
\multirow{5}{*}{\texttt{gpt-5.5}} & B0$\cdot$closed & 0.0519 & 0.0593 & 0.0296 & 0.0259 & 0.0149 & B0$\cdot$open & 0.9370 & 0.7333 & 0.7370 & 0.5704 & 0.6320 \\
 & M1$\cdot$closed & 0.0630 & 0.0704 & 0.0519 & 0.0593 & 0.0520 & M2$\cdot$active & 0.9296 & 0.7444 & 0.7593 & 0.6037 & 0.6952 \\
 & cascade$\cdot$open & 0.9444 & 0.7519 & 0.7296 & 0.5926 & 0.6729 & M3$\cdot$closed & 0.0407 & 0.0481 & 0.0259 & 0.0370 & 0.0149 \\
 & M3$\cdot$open & 0.9370 & 0.7370 & 0.7259 & 0.5741 & 0.6171 & M4$\cdot$closed & 0.0481 & 0.0444 & 0.0185 & 0.0296 & 0.0260 \\
 & M4$\cdot$open & 0.9407 & 0.7481 & 0.7481 & 0.6037 & 0.6654 & N/A & \multicolumn{5}{c}{N/A} \\
\midrule
\multirow{5}{*}{\texttt{qwen3-235b}} & B0$\cdot$closed & 0.0148 & 0.0259 & 0.0037 & 0.0037 & 0.0037 & B0$\cdot$open & 0.9370 & 0.7444 & 0.7222 & 0.5630 & 0.6245 \\
 & M1$\cdot$closed & 0.0111 & 0.0000 & 0.0000 & 0.0000 & 0.0000 & M2$\cdot$active & 0.8630 & 0.7259 & 0.7148 & 0.5852 & 0.6729 \\
 & cascade$\cdot$open & 0.8556 & 0.7333 & 0.7037 & 0.5852 & 0.6617 & M3$\cdot$closed & 0.0037 & 0.0074 & 0.0074 & 0.0037 & 0.0074 \\
 & M3$\cdot$open & 0.8778 & 0.7148 & 0.7148 & 0.5481 & 0.6208 & M4$\cdot$closed & 0.0037 & 0.0000 & 0.0000 & 0.0037 & 0.0000 \\
 & M4$\cdot$open & 0.8963 & 0.7296 & 0.7185 & 0.6037 & 0.6766 & N/A & \multicolumn{5}{c}{N/A} \\
\midrule
\multirow{5}{*}{\texttt{gemini-3.1-pro}} & B0$\cdot$closed & 0.1444 & 0.0926 & 0.0444 & 0.0815 & 0.0558 & B0$\cdot$open & 0.9185 & 0.7593 & 0.7444 & 0.5741 & 0.6171 \\
 & M1$\cdot$closed & 0.1407 & 0.1407 & 0.0593 & 0.1185 & 0.1561 & M2$\cdot$active & 0.9222 & 0.7519 & 0.6889 & 0.5296 & 0.6543 \\
 & cascade$\cdot$open & 0.9074 & 0.7296 & 0.6926 & 0.5148 & 0.6468 & M3$\cdot$closed & 0.1370 & 0.0815 & 0.0481 & 0.0963 & 0.0706 \\
 & M3$\cdot$open & 0.9037 & 0.7556 & 0.7222 & 0.5407 & 0.6134 & M4$\cdot$closed & 0.1741 & 0.1185 & 0.1000 & 0.1259 & 0.1413 \\
 & M4$\cdot$open & 0.9519 & 0.7593 & 0.7259 & 0.5593 & 0.6543 & N/A & \multicolumn{5}{c}{N/A} \\
\midrule
\multirow{5}{*}{\texttt{glm-5.2}} & B0$\cdot$closed & 0.0222 & 0.0259 & 0.0074 & 0.0037 & 0.0037 & B0$\cdot$open & 0.9333 & 0.7074 & 0.7222 & 0.5556 & 0.6059 \\
 & M1$\cdot$closed & 0.0074 & 0.0037 & 0.0037 & 0.0037 & 0.0000 & M2$\cdot$active & 0.9370 & 0.7481 & 0.7519 & 0.6259 & 0.7398 \\
 & cascade$\cdot$open & 0.9259 & 0.7111 & 0.7148 & 0.5704 & 0.6134 & M3$\cdot$closed & 0.0000 & 0.0111 & 0.0037 & 0.0037 & 0.0074 \\
 & M3$\cdot$open & 0.8185 & 0.6185 & 0.6111 & 0.4593 & 0.4870 & M4$\cdot$closed & 0.0037 & 0.0000 & 0.0037 & 0.0037 & 0.0037 \\
 & M4$\cdot$open & 0.9259 & 0.7222 & 0.7037 & 0.5704 & 0.6357 & N/A & \multicolumn{5}{c}{N/A} \\
\bottomrule
\end{tabular}
\caption{Combined layout of split-wise glm-5.2 results and difficulty-wise results for all five models. Panel (a) retains L1--L4 for glm-5.2 on four splits and nine conditions. Panel (b) pairs the nine conditions for each model and lists L4 accuracy from D1 through D5. The right side of the final row is marked N/A because the number of conditions is odd. D1--D4 each contain $n=270$ samples, and D5 contains $n=269$.}
\label{tab:persplit-glm52}
\label{tab:perdiff}
\end{table*}

\begin{table*}[p]
\centering
\small
\setlength{\tabcolsep}{3pt}
\begin{tabular}{lrrrrrrrr}
\multicolumn{9}{c}{\textbf{(a) M2 Active Retrieval: Cost, Submission Quality, and Tool Use}} \\
\toprule
model & retrievals & rounds & reach L4 & full-EC precision & BLAST & HMMER & structure & active site \\
\midrule
deepseek-v4-pro & 1.667 & 5.265 & 0.8028 & 0.8873 & 1257 & 624 & 339 & 29 \\
gpt-5.5         & 1.325 & 5.180 & 0.8547 & 0.8734 & 1260 & 363 & 145 & 19 \\
qwen3-235b      & 1.678 & 5.468 & 0.8695 & 0.8193 & 1190 & 537 & 487 & 50 \\
gemini-3.1-pro  & 1.636 & 5.322 & 0.8310 & 0.8537 & 1265 & 615 & 315 & 12 \\
glm-5.2         & 1.477 & 5.371 & 0.9014 & 0.8438 & 1266 & 543 & 153 & 31 \\
\bottomrule
\end{tabular}

\setlength{\tabcolsep}{4.2pt}
\begin{tabular}{llrrrr}
\multicolumn{6}{c}{\textbf{(b) M4 Self-Consistency Diagnostics in Closed- and Open-Book Settings}} \\
\toprule
model & setting & consensus depth & winning share & reach L4 & full-EC precision \\
\midrule
\multirow{2}{*}{deepseek-v4-pro} & closed & 0.756 & 0.772 & 0.0600 & 0.7284 \\
 & open & 3.591 & 0.979 & 0.8132 & 0.9015 \\
\multirow{2}{*}{gpt-5.5} & closed & 1.652 & 0.759 & 0.1757 & 0.1899 \\
 & open & 3.670 & 0.986 & 0.8228 & 0.9009 \\
\multirow{2}{*}{qwen3-235b} & closed & 0.501 & 0.726 & 0.0082 & 0.1818 \\
 & open & 3.591 & 0.933 & 0.8251 & 0.8787 \\
\multirow{2}{*}{gemini-3.1-pro} & closed & 1.844 & 0.790 & 0.2854 & 0.4623 \\
 & open & 3.636 & 0.973 & 0.7932 & 0.9206 \\
\multirow{2}{*}{glm-5.2} & closed & 0.803 & 0.719 & 0.0074 & 0.4000 \\
 & open & 3.569 & 0.958 & 0.7769 & 0.9160 \\
\bottomrule
\end{tabular}
\setlength{\tabcolsep}{4.2pt}
\begin{tabular}{lccc}
\multicolumn{4}{c}{\textbf{(c) Bootstrap Confidence Intervals for L4 Accuracy}} \\
\toprule
model & B0 closed L4 [95\% CI] & B0 open L4 [95\% CI] & best condition L4 [95\% CI] \\
\midrule
deepseek-v4-pro & 0.159 [0.140, 0.181] & 0.731 [0.707, 0.756] & 0.733 [0.709, 0.758] (M4 open) \\
gpt-5.5         & 0.036 [0.026, 0.047] & 0.722 [0.698, 0.748] & 0.747 [0.724, 0.771] (M2) \\
qwen3-235b      & 0.010 [0.006, 0.016] & 0.718 [0.694, 0.744] & 0.725 [0.701, 0.749] (M4 open) \\
gemini-3.1-pro  & 0.084 [0.070, 0.098] & 0.723 [0.698, 0.747] & 0.730 [0.706, 0.756] (M4 open) \\
glm-5.2         & 0.013 [0.007, 0.019] & 0.705 [0.681, 0.731] & 0.761 [0.738, 0.783] (M2) \\
\bottomrule
\end{tabular}
\par\vspace{2pt}
\setlength{\tabcolsep}{5pt}
\begin{tabular}{lrrrrr}
\multicolumn{6}{c}{\textbf{(d) Exact Values by Maximum Sequence Homology}} \\
\toprule
homology bin & $<30$ & 30--40 & 40--50 & 50--70 & $\geq 70$ \\
\midrule
$n$              & 179  & 380  & 460  & 198  & 132 \\
open-book L4     & 0.2849 & 0.6268 & 0.8422 & 0.8808 & 0.9091 \\
closed-book L4   & 0.0302 & 0.0421 & 0.0643 & 0.0848 & 0.1045 \\
open minus closed & 0.2547 & 0.5847 & 0.7779 & 0.7960 & 0.8046 \\
\bottomrule
\end{tabular}
\caption{Complete diagnostics for active retrieval, self-consistency, interval estimates, and homology bins. In panel~(a), \emph{retrievals} is the mean number of tool calls per enzyme and \emph{rounds} is the mean number of model turns per enzyme, both over all $1349$ items, and the last four columns count tool \emph{calls} rather than samples, so they sum to $1349$ times the retrieval column. \texttt{esm\_knn} is listed in the tool menu whenever the cache holds neighbors for the query and is still never chosen by any model, whereas the \texttt{pfam} channel is registered in the toolbox but was never populated, so it is never offered and never called; both are therefore zero. Full-EC precision uses samples with a complete four-level EC prediction as the denominator. Panel (c) uses 1000 sample-level bootstrap replicates. Panel~(d) pools the five models under B0, so it reproduces the homology curve of the main text; the counts sum to 1349 and homology is the maximum local BLAST identity in percent.}
\label{tab:full-diagnostics}
\label{tab:homology-bins}
\end{table*}

\begin{table*}[p]
\centering
\small
\setlength{\tabcolsep}{4.2pt}
\begin{tabular}{lccccccccccc}
\toprule
& \multicolumn{5}{c}{M2 active retrieval} && \multicolumn{5}{c}{M4$\cdot$open} \\
\cmidrule(lr){2-6}\cmidrule(lr){8-12}
model & \texttt{<30\%} & \texttt{30--50\%} & \texttt{price} & \texttt{promis.} & full && \texttt{<30\%} & \texttt{30--50\%} & \texttt{price} & \texttt{promis.} & full \\
\midrule
deepseek-v4-pro & 0.634 & 0.830 & 0.439 & 0.751 & 0.712 && 0.620 & 0.836 & 0.460 & 0.885 & 0.733 \\
gpt-5.5         & 0.639 & 0.830 & 0.514 & 0.909 & 0.747 && 0.641 & 0.832 & 0.439 & 0.919 & 0.741 \\
qwen3-235b      & 0.637 & 0.782 & 0.493 & 0.837 & 0.712 && 0.613 & 0.804 & 0.473 & 0.923 & 0.725 \\
gemini-3.1-pro  & 0.620 & 0.813 & 0.487 & 0.775 & 0.709 && 0.634 & 0.838 & 0.385 & 0.885 & 0.730 \\
glm-5.2         & 0.639 & 0.830 & 0.561 & 0.967 & 0.761 && 0.595 & 0.805 & 0.405 & 0.919 & 0.712 \\
\midrule
five-model mean & 0.634 & 0.817 & 0.499 & 0.848 & 0.728 && 0.621 & 0.823 & 0.432 & 0.906 & 0.728 \\
neighbor vote   & 0.618 & 0.816 & 0.372 & 0.938 & 0.723 && 0.618 & 0.816 & 0.372 & 0.938 & 0.723 \\
$\Delta$ vs vote & $+0.016$ & $+0.001$ & $+0.127$ & $-0.090$ & $+0.005$ && $+0.003$ & $+0.007$ & $+0.060$ & $-0.032$ & $+0.005$ \\
\bottomrule
\end{tabular}
\caption{L4 accuracy of the two best LLM settings against plain neighbor voting, per model and per split. Neighbor vote is the \texttt{BLAST plus HMMER} row of panel (b) of Table~\ref{tab:baselines} and involves no LLM. Split sizes are $432$, $560$, $148$ and $209$. The two LLM settings are indistinguishable on the full set and disagree by $0.067$ on \texttt{price} and $0.058$ on \texttt{promiscuous}.}
\label{tab:netgain-full}
\end{table*}